\documentclass[10pt,journal,compsoc]{IEEEtran}
\usepackage{setspace}

\ifCLASSOPTIONcompsoc
  \usepackage[nocompress]{cite}
\else
  \usepackage{cite}
\fi

\usepackage{amsmath,amssymb}
\usepackage[T1]{fontenc}
\usepackage{multirow}
\usepackage{booktabs}
\ifCLASSINFOpdf
  \usepackage[pdftex]{graphicx}
\else
\fi
\usepackage{algorithmic}
\begin{document}
%
\title{On the Evolution of Knowledge Graphs: A Survey and Perspective}

\author{Xuhui Jiang$^{\dagger}$, Chengjin Xu$^{\dagger}$, Yinghan Shen, Xun Sun, Lumingyuan Tang, Saizhuo Wang, \\ Zhichao Shi, Baozhu Shang, Zhongwu Chen, Yuanzhuo Wang, Jian Guo$^{*}$
\IEEEcompsocitemizethanks{\IEEEcompsocthanksitem Xuhui Jiang was with the International Digital Economy Academy (IDEA), and Institute of Computing Technology, Chinese Academy of Sciences, China.\protect\\
E-mail: jiangxuhui19g@ict.ac.cn
\IEEEcompsocthanksitem Both Xuhui Jiang and Chengjin Xu contribute equally.
\IEEEcompsocthanksitem Jian Guo is the corresponding author.
}

}

\IEEEtitleabstractindextext{%
\begin{abstract}
Knowledge graphs (KGs) are structured representations of diversified knowledge. They are widely used in various artificial intelligence applications.
In this article, we provide a comprehensive survey on the evolution of various types of knowledge graphs (i.e., static KGs, dynamic KGs, temporal KGs, and event KGs) and techniques for knowledge extraction and reasoning.
Furthermore, we introduce the practical applications of different types of KGs, including a case study in financial analysis. 
Finally, we propose our perspective on the future directions of knowledge engineering, including the potential of combining the power of knowledge graphs and large language models (LLMs), and the evolution of knowledge extraction, reasoning, and representation.
\end{abstract}
}

\maketitle

\IEEEdisplaynontitleabstractindextext

%
\IEEEpeerreviewmaketitle

\IEEEraisesectionheading{\section{Introduction}\label{sec:introduction}}

\IEEEPARstart{K}{nowledge} graph (KG) is a widely-used knowledge representation technology in artificial intelligence research. It can organize huge amounts of scattered data into structured knowledge. 
Typical KGs including Google Knowledge Graph~\cite{googleKG}, DBpedia~\cite{dbpedia}, Freebase~\cite{freebase} and YAGO~\cite{YAGO}, have been widely used in a variety of intelligent applications, such as search engines~\cite{ehrlinger2016survey}, question answering systems~\cite{ji2020survey}, recommendation systems~\cite{survey1}, etc.

\begin{figure}[t]
	\centering
	\includegraphics[width=1\columnwidth]{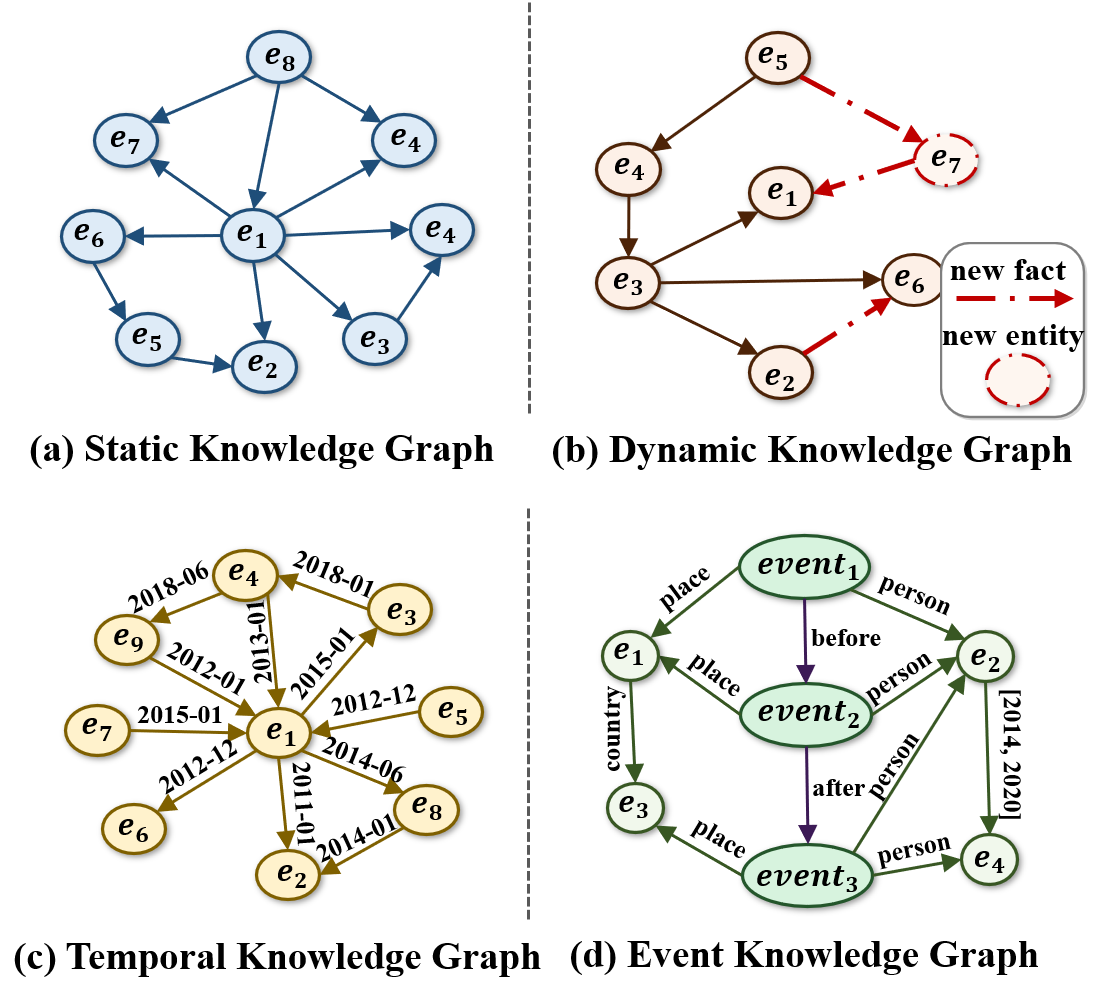}
	\caption{Examples of four categories of the knowledge graph, i.e., static, dynamic, temporal, and event knowledge graph.}
	\label{fig:fig_kgs}
\end{figure}

\begin{figure*}[t]
	\centering
	\includegraphics[width=2\columnwidth]{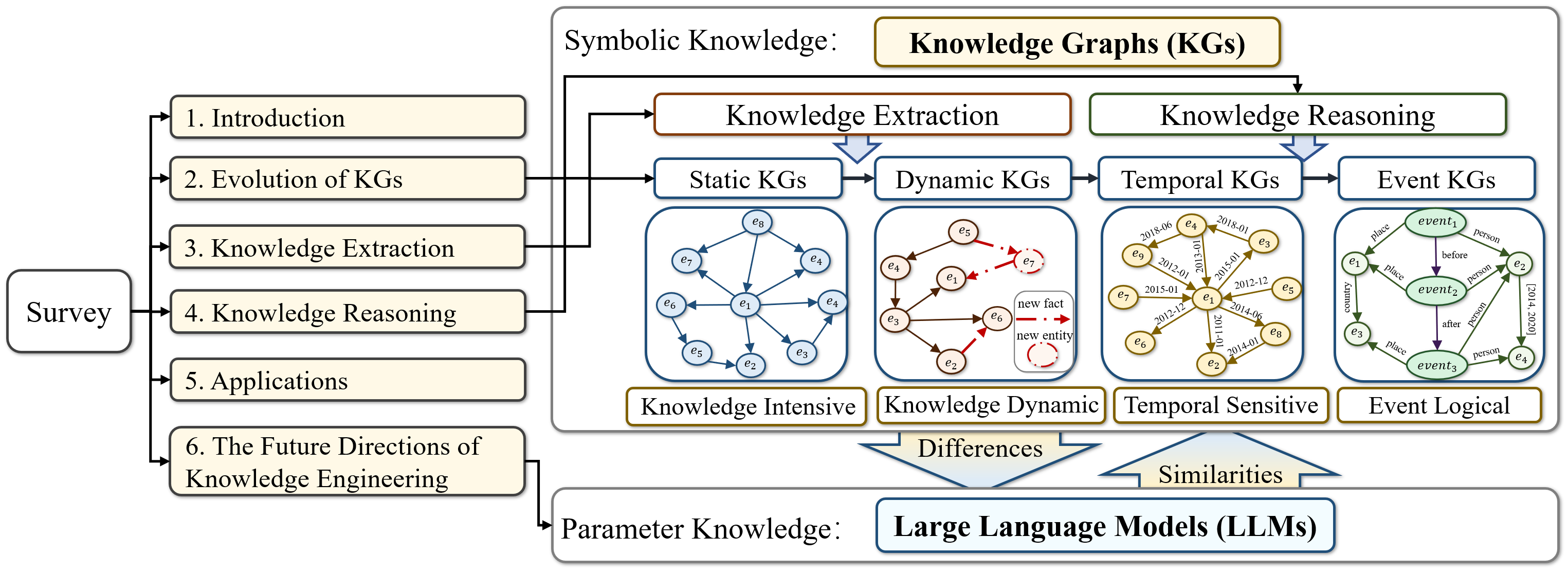}
	\caption{General framework of the survey.}
	\label{fig:fig_sub}
\end{figure*}

A Knowledge Graph (KG) is depicted as a graph where nodes represent entities that describe either real-world objects or abstract concepts, and edges denote relations that signify semantic or logical connections between these entities.
Therefore, the basic building blocks of a KG are facts shaped like (\textit{subject entity}, \textit{relation}, \textit{object entity}), which are also known as triples ($e_s$, $r$, $e_o$).

Although the form of fact triples facilitates the representation of static knowledge, most real-world knowledge elements are always evolving or changing. Thus, static KGs (SKGs) where triples are never updated or updated very infrequently, are limited due to the timeliness of the knowledge they hold~\cite{ATiSE}. 
To solve this problem, many dynamic KGs (DKGs) continuously collect and update knowledge to ensure the timeliness of knowledge. Moreover, temporal KGs (TKGs) such as ICEWS~\cite{ICEWS}, GDELT~\cite{GDELT}, YAGO3~\cite{YAGO3} and Wikidata~\cite{Wikidata} which attach time information into triple facts have drawn more and more interest recently.
Besides entity-centric knowledge, events are also an essential kind of knowledge in the world, which trigger the spring up of event-centric knowledge representation form such as Event KG (EKG)~\cite{event-centric}. Noteworthy, events are dynamic and temporal in nature and relations between events and facts are often temporal or causal ones. The four different categories of KG are shown in Figure~\ref{fig:fig_kgs}.

A vast amount of knowledge exists in unstructured data sources.
Knowledge extraction (KGE), which focuses on automatically extracting structured information from unstructured data, is one of the key technologies of knowledge graph construction. Knowledge extraction for different types of KGs involves various NLP techniques, including named entity recognition (NER)~\cite{ner_}, relation extraction~\cite{REsurvey1}, event extraction and rule extraction.
These knowledge extraction techniques developed rapidly in recent years with the rise of large language models (LLMs), and they are extensively reviewed in this paper.

In addition, most KGs reviewed in this paper are suffering from the problem of incompleteness, and it severely limits the metric of KG application to downstream tasks. Thus, KG reasoning (KGR), a task that aims to predict and complete missing facts in KGs, has become one of the research hot spots in KG studies. With the development of KGs, KGR techniques have also evolved from static learning methods to temporal learning methods. According to the types of target KGs, we classify the existing KGR methods into three categories, i.e., static KGR (SKGR) methods, dynamic KGR (DKGR) methods and temporal KGR (TKGR) methods. We will discuss event KGR (EKGR) as a part of future work in this paper, since this task needs more research in the future. 


Although there are several survey papers focusing on knowledge graphs and the relevant techniques of knowledge extraction and KGR, they each only focus on technologies related to a single type of KGs and thus do not study the difference and connection between the types of their target KGs.  
The goal of this survey is to provide a comprehensive and up-to-date overview of KGs from an evolving perspective (i.e., SKG$\rightarrow$DKG$\rightarrow$TKG$\rightarrow$EKG), and highlight promising future research directions. Our contributions can be summarized as follows:

\begin{itemize}
    \item \textbf{Comprehensive Review.} We study the various KGs from an evolution perspective, and provide a comprehensive review of knowledge extraction and reasoning techniques for different types of KGs. 
    \item \textbf{New perspectives and insights on future direction.} We introduce our perspectives in the future directions of knowledge engineering, especially the connection to LLMs.
\end{itemize}

The article is organized and shown in Figure~\ref{fig:fig_sub}.
Sec.~\ref{sec:introduction} introduces the formal evolution of KGs, as well as relevant definitions and notations; Sec.~\ref{sec:3} and Sec.~\ref{sec:4} review the knowledge extraction and reasoning techniques applied to different types of KGs; Sec.~\ref{sec:5} illustrates a case study of KGs in financial analysis and other applications;
Sec.~\ref{sec:6} discusses future research directions about knowledge engineering.

\begin{figure*}[t]
	\centering
	\includegraphics[width=1.7\columnwidth]{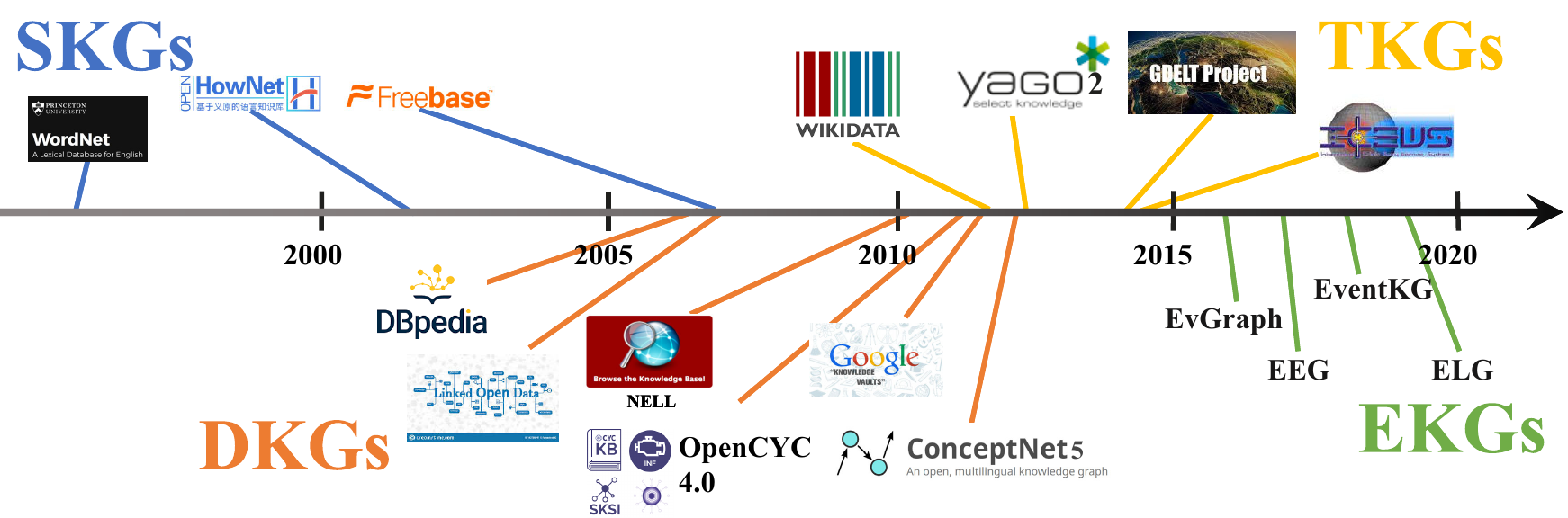}
	\caption{The formal evolution history of KGs.}
	\label{fig:kg_evo}
\end{figure*}

\section{Evolution of Knowledge Graphs}\label{sec:preliminary}
The development of knowledge graphs can be traced back to the birth of expert systems in the last century. After that, the concept of knowledge graphs was gradually formed after the development of the World Wide Web 1.0 era and the Semantic Web. Starting in 2006, the emergence
of large-scale Wikipedia-like rich-structure knowledge resources and advances in web-scale information extraction methods have led to tremendous progress in knowledge extraction methods, which are automated and operated at the web scale. Currently, automatically-built KGs have become powerful assets for semantic search, big data analysis, intelligent recommendations, and data integration, and are being widely used in large industries and various domains.

Early KGs were mainly used to store static knowledge. As the increasing demand of downstream tasks to the timeliness of knowledge, KGs that can continuously update knowledge or store time information of facts have emerged. Nevertheless, entity-centric KGs are still unable to elegantly represent the event information that appears in news texts. Thus, event-centric KGs have been proposed, and become popular in specific domains, such as finance and tourism.

Based on this observation, we divide the formal evolution of KGs into four stages, corresponding to the emergence of static KGs (SKGs), dynamic KGs (DKGs), the temporal KGs (KGs) stage, and event KGs (EKGs), respectively. In this section, we first summarize the notation as shown in Table~\ref{table:notation}, then we will provide definitions of the four types of KGs, and their typical representatives, as well as their application scenarios.

\begin{table}
\centering
\caption{Overview of Knowledge Graph Reasoning Methods}
\label{table:notation}
\begin{tabular}{cc}
\toprule
Notation & Explanation \\
\midrule
$\mathcal{SKG}$ & Static knowledge graph\\
$\mathcal{DKG}$ & Dynamic knowledge graph\\
$\mathcal{TKG}$ & Temporal knowledge graph \\
$\mathcal{EKG}$ & Event knowledge graph \\
$\mathcal{E}_n$ & Entity set\\
$\mathcal{E}_v$ & Event set\\
$\mathcal{R}$ & Relation set\\
$\mathcal{F}$ & The set of facts\\
$\mathcal{T}$ & The set of the time stamps\\
$\mathcal{F}_t$ & The set of facts at time $t$ \\
$\left(e_s, r_p, e_o\right)$ & Fact of the head, relation, tail.\\
$\left(e_s, r_p, e_o, t\right)$ & Fact of the head, relation, tail, timestamp\\
\bottomrule
\end{tabular}
\end{table}

\subsection{Static Knowledge Graphs}
\begin{figure}[h]
	\centering
	\includegraphics[width=0.8\columnwidth]{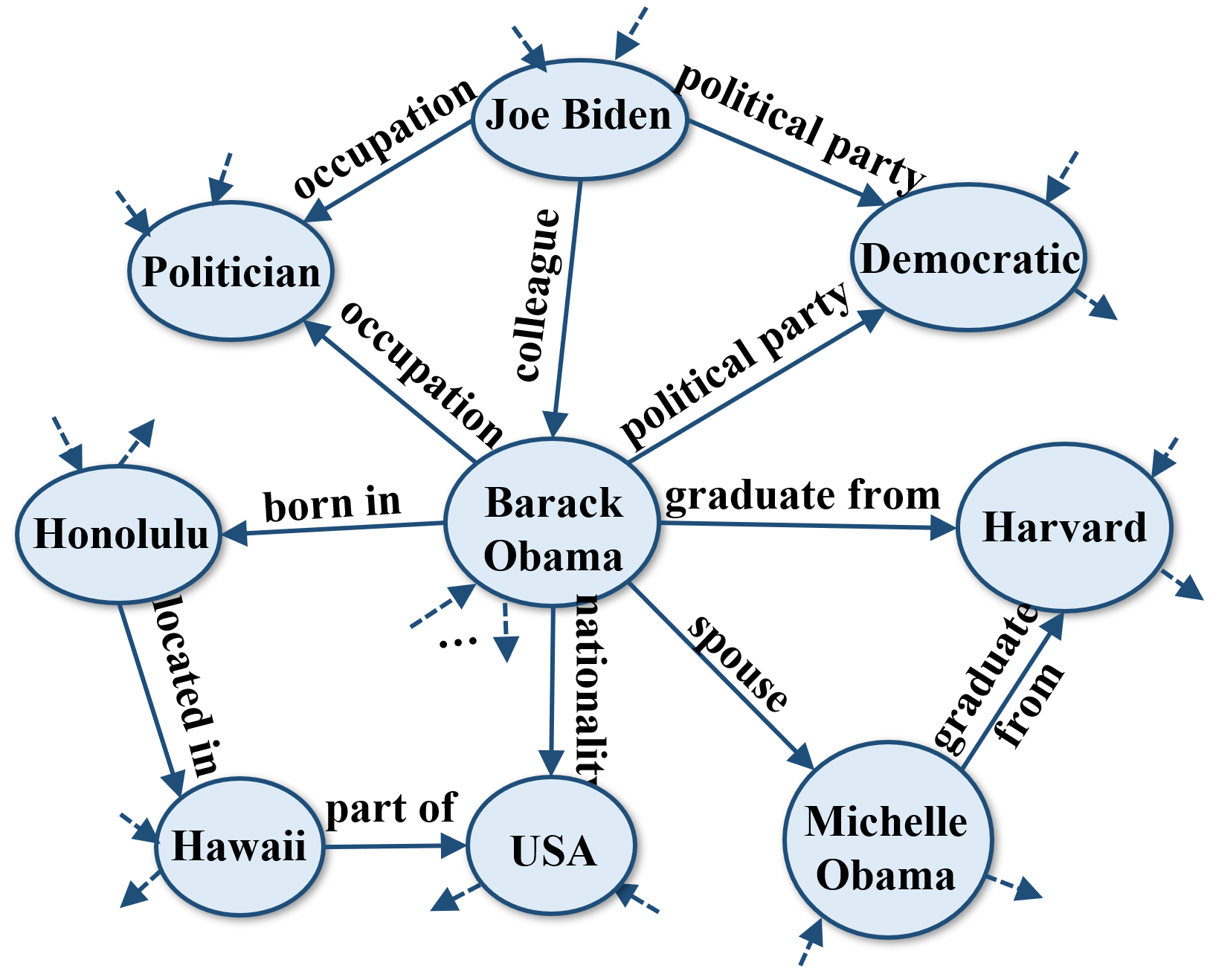}
	\caption{Examples of a static knowledge graph.}
	\label{fig:fig_skg}
\end{figure}

The first generation of KGs is static graphical representations that present knowledge in the form of entities and relations between entities, as shown in Figure~\ref{fig:fig_skg}. The formal definition is declared as follows,
\par\smallskip

\noindent\textbf{Definition 1. Static Knowledge Graph.}
~\textit{
A Static Knowledge Graph is defined as $\mathcal{SKG}=\{\mathcal{E}_{n},\mathcal{R},\mathcal{F}\}$, where $\mathcal{E}_{n}$ and $\mathcal{R}$ represent the sets of entities and relations, respectively. $\mathcal{F}\subseteq\mathcal{E}_{n}\times\mathcal{R}\times\mathcal{E}_{n}$ is a set of facts $\{(e_{s},r_{p},e_{o})\}$, where $e_{s}, e_{o}\in\mathcal{E}_{n}$ and $r_{p}\in\mathcal{R}$. A fact ${(e_{s},r_{p},e_{o})}$ represents that relation $r_p$ exists between subject entity $e_s$ and object entity $e_{o}$ at unspecified time.
}

A few examples of SKGs include Kinship~\cite{kinship}, HowNet~\cite{dong2003hownet}, WordNet~\cite{wordnet}, and Freebase~\cite{freebase}. 
Kinship database is a small-size relational database, consisting of 24 unique names in two families. HowNet is a semantic knowledge base, designed to capture the meaning of words and concepts in Chinese. These two KGs are static representations of relationships between specific entities and concepts, and are not updated after creation.
WordNet is a lexical database of English words and concepts, which includes synonyms and hypernyms. 
Freebase was designed to be an open and shared repository of structured data about a wide range of topics, including people, places, organizations, and more. Since they have no longer been updated for a long time, they can be regarded as SKGs.

SKGs are useful for many applications, such as information retrieval and question answering, where a fixed set of relations between entities is required. However, they are not suitable for applications that require up-to-date information, such as news articles or social media feeds.

\subsection{Dynamic Knowledge Graphs}
\begin{figure}[h]
	\centering
	\includegraphics[width=0.8\columnwidth]{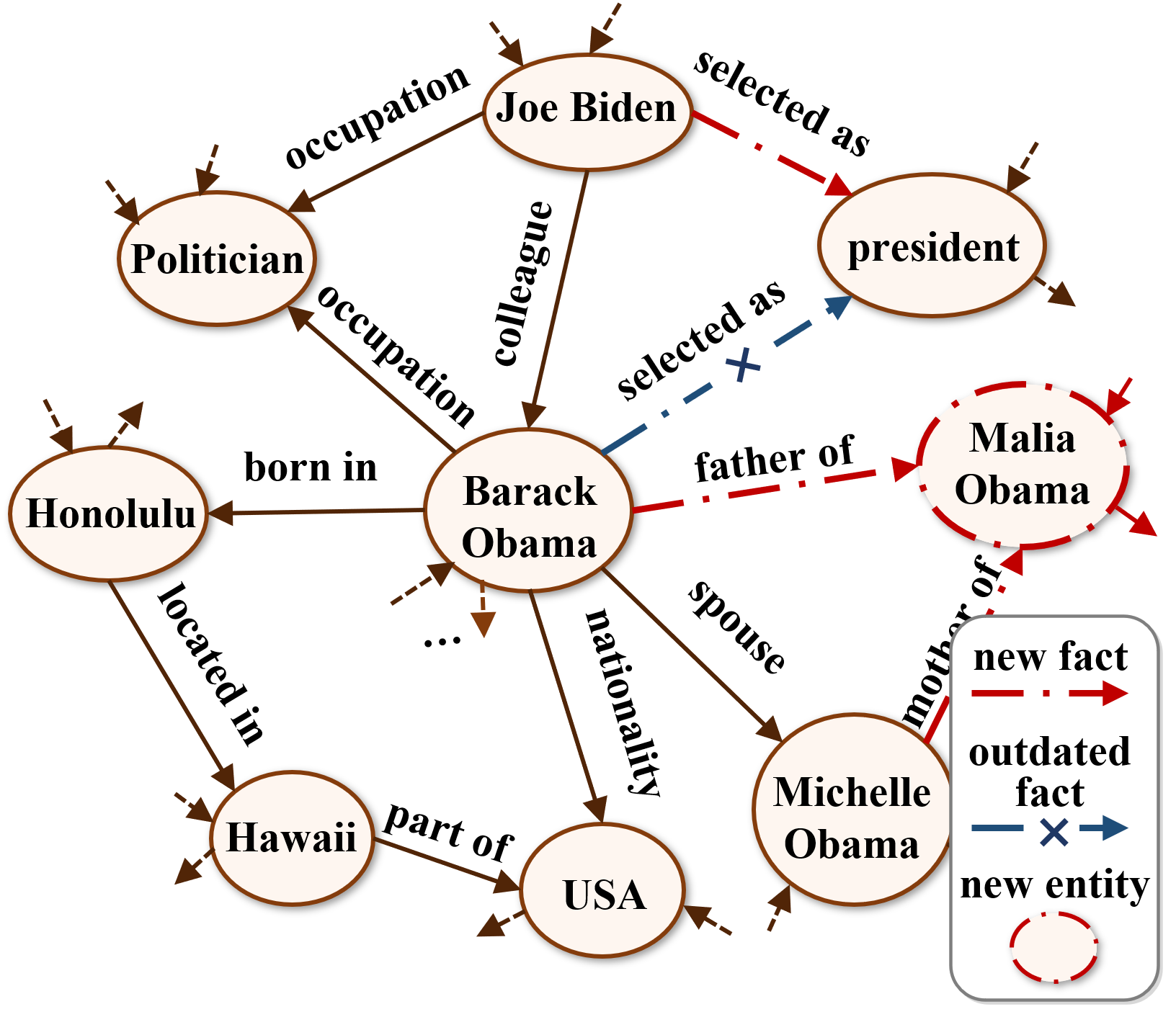}
	\caption{Examples of a dynamic knowledge graph.}
	\label{fig:fig_dkg}
\end{figure}

The second generation of KGs introduced the concept of change and updates to the graph, guaranteeing the timeliness of knowledge and enabling the extensibility of the graph, as shown in Figure~\ref{fig:fig_dkg}. The formal definition is declared as follows,
\par\smallskip

\noindent\textbf{Definition 2. Dynamic Knowledge Graph.}
~\textit{
A Dynamic Knowledge Graph is defined as $\mathcal{DKG}=\{\mathcal{E}_{n},\mathcal{R},\mathcal{F}\}$, where $\mathcal{E}_{n}$ and $\mathcal{R}$ represent the dynamic sets of entities and relations, respectively. $\mathcal{F}\subseteq\mathcal{E}_{n}\times\mathcal{R}\times\mathcal{E}_{n}$ is a dynamic set of facts $\{(e_{s},r_{p},e_{o})\}$, where $e_{s}, e_{o}\in\mathcal{E}_{n}$ and $r_{p}\in\mathcal{R}$. A fact ${(e_{s},r_{p},e_{o})}$ represents that relation $r_p$ exists between subject entity $e_s$ and object entity $e_{o}$ at the time of the latest update.
}

Most KGs are dynamic and evolving. For example, DBpedia~\cite{dbpedia} is a DKG that is created by extracting structured data from Wikipedia~\cite{wikipedia}. It is updated regularly to reflect changes in Wikipedia. Similarly, many large-scale KGs like Google Knowledge Graph~\cite{googleKG}, NELL~\cite{nell}, Linked Open Data~\cite{LinkedOpenData}, OpenCyc~\cite{Opencyc}, ConceptNet~\cite{conceptnet} and UMLS~\cite{UMLS}, continuously updated over time, even in real-time, to reflect changes in the underlying data. These graphs are designed to be more flexible and adaptable than SKGs, allowing them to better handle data that is constantly changing and evolving. It is worth noting that a separate version of a DKG can be viewed as an SKG. And a DKG that is no longer updated will also degenerate into an SKG. Therefore, an SKG can be viewed as a special case of a DKG, i.e., \textbf{Definition 2} subsumes \textbf{Definition 1}.

DKGs are being used in various applications, including search engines, recommendation systems, chatbots, and more. By continuously updating the KGs with the latest data, these systems can provide users more accurate and relevant results. However, DKGs can not be applied in some applications such as historical analysis, trend analysis and forecasting, due to the loss of historical data.

\subsection{Temporal Knowledge Graphs}
\begin{figure}[h]
	\centering
	\includegraphics[width=0.9\columnwidth]{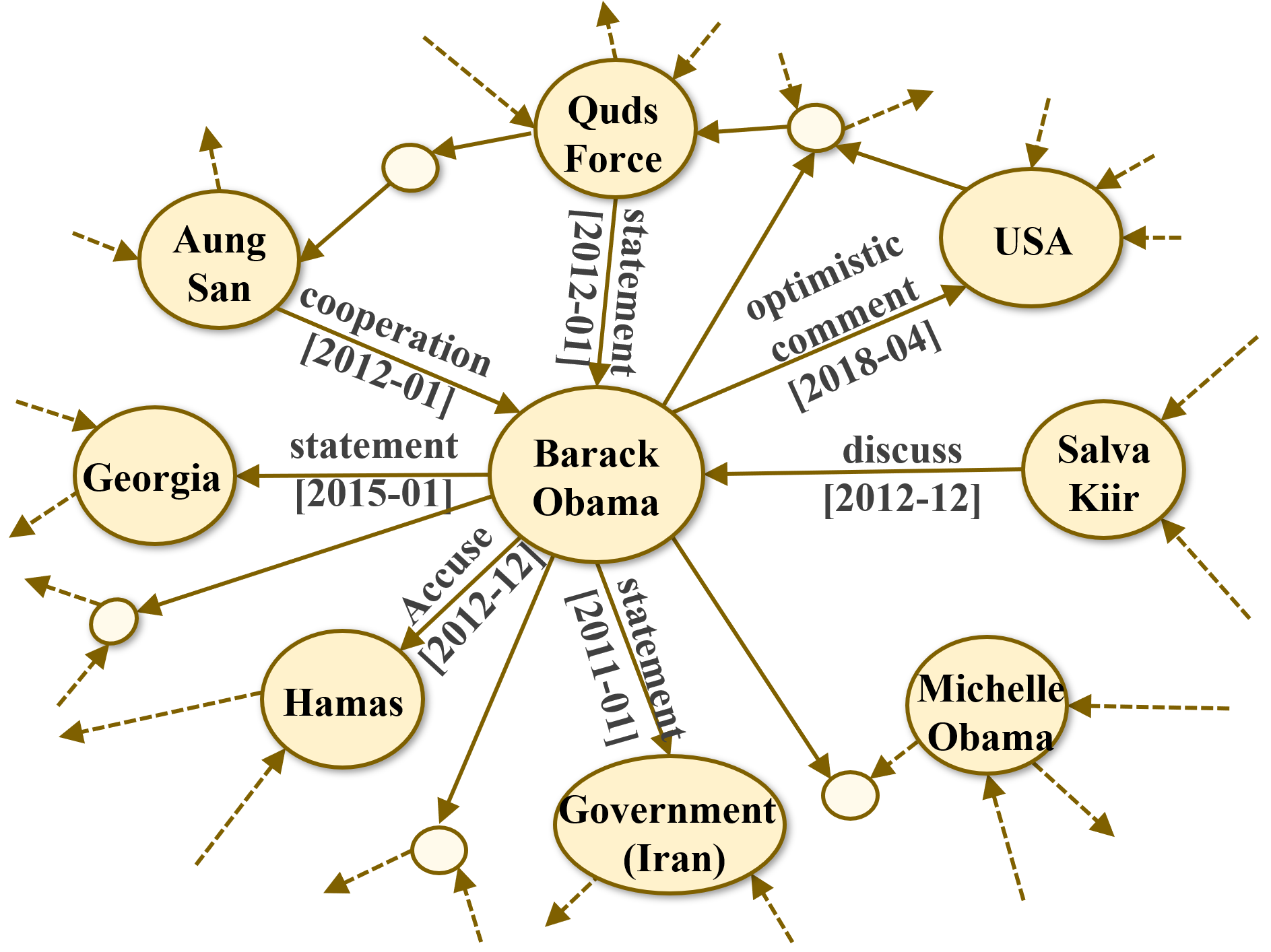}
	\caption{Examples of a temporal knowledge graph.}
	\label{fig:fig_tkg}
\end{figure}

The third generation of KGs adds a temporal dimension to the knowledge, allowing for the representation of time-varying relations between entities. The temporal knowledge graphs are designed to capture and represent knowledge that changes over time, and tend to be dynamic in nature with their time-sensitivity, as shown in Figure~\ref{fig:fig_tkg}. The formal definition is declared as follows,
\par\smallskip

\noindent\textbf{Definition 3. Temporal Knowledge Graph.}
~\textit{
A Temporal Knowledge Graph is defined as $\mathcal{TKG}=\{\mathcal{E}_{n},\mathcal{R},\mathcal{T},\mathcal{F}\}$, where $\mathcal{E}_{n}$, $\mathcal{R}$ and $\mathcal{T}$ represent the extendable sets of entities, relations and timestamps, respectively. $\mathcal{F}\subseteq\mathcal{E}_{n}\times\mathcal{R}\times\mathcal{E}_{n}\times\mathcal{T}$ is an extendable set of facts $\{(e_{s},r_{p},e_{o},t)\}$, where $e_{s}, e_{o}\in\mathcal{E}_{n}$, $r_{p}\in\mathcal{R}$ and $t\in\mathcal{T}$. A fact ${(e_{s},r_{p},e_{o},t)}$ represents that relation $r_p$ exists between subject entity $e_s$ and object entity $e_{o}$ at time $t$. $t$ is sometimes unknown and can be represented in different forms, e.g., time points or time intervals.
}

SKGs and DKGs represent facts in the form of triples, shaped like (\textit{Donald Trump}, \textit{president of}, \textit{USA}). The time spans when facts hold are unknown. Thus, one can not use SKGs or DKGs to tackle time-sensitive tasks, like time-aware link prediction, forecasting and others. A lot of large-scale TKGs like Wikidata~\cite{Wikidata}, YAGO~\cite{YAGO3} associate fact triples with valid time
annotations.
Some event databases, e.g., ICEWS~\cite{ICEWS} and GDELT~\cite{GDELT} can also be taken as TKGs since they decompose event news into fact quadruples and store them. Such TKGs provide rich temporal information, but not every fact in a TKG includes a time stamp. Thus, DKGs can be regarded as special cases of TKGs, where all-time information is unobtainable. In another words, \textbf{Definition 3} subsumes \textbf{Definition 2}.

Overall, TKGs offer a more comprehensive view of data over time, which is useful for a wide range of applications, including temporal query answering, and predictive analysis. On the other hand, some TKGs are extracted from event news, but do not effectively represent event information due to their entity-centric representations.

\subsection{Event Knowledge Graphs}
\begin{figure}[h]
	\centering
	\includegraphics[width=0.8\columnwidth]{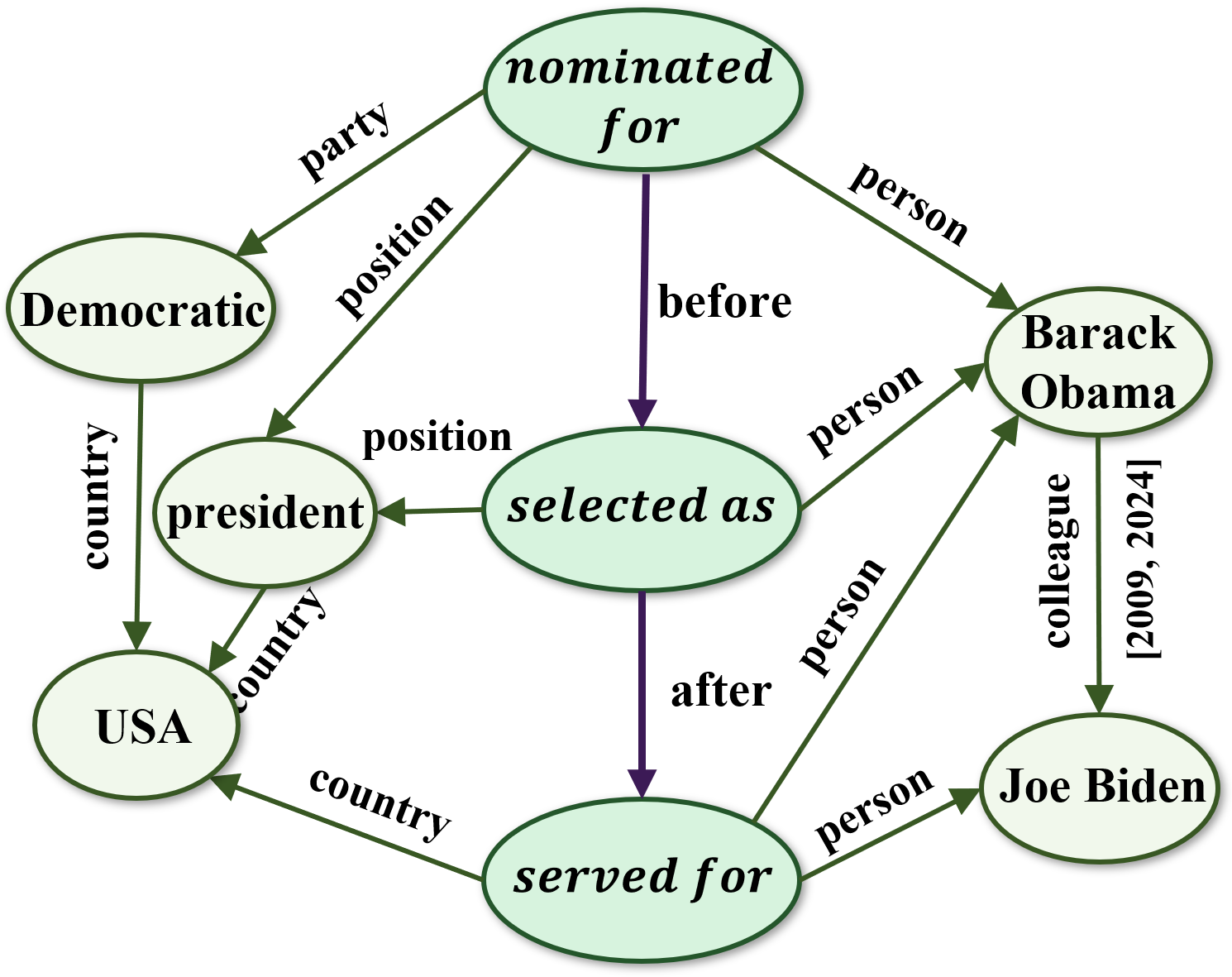}
	\caption{Examples of an event knowledge graph.}
	\label{fig:fig_ekg}
\end{figure}

The most recent generation of KGs focuses on representing and understanding events, which are time-bound occurrences that involve multiple entities and relations, enabling the representation of complex, time-varying relations between entities and events, as shown in Figure~\ref{fig:fig_ekg}. The formal definition is declared as follows,
\par\smallskip

\noindent\textbf{Definition 4. Event Knowledge Graph.}
~\textit{
An Event Knowledge Graph is defined as $\mathcal{EKG}=\{\mathcal{E},\mathcal{R},\mathcal{T},\mathcal{F}\}$. $\mathcal{E}=\mathcal{E}_{n}\cup\mathcal{E}_{v}$ is the node set, where $\mathcal{E}_{v}$ is an extendable event set. $\mathcal{R}$ represents the sets of relations between entities and events. $\mathcal{F}\subseteq\mathcal{E}\times\mathcal{R}\times\mathcal{E}\times\mathcal{T}$ is an extendable set of facts $\{(e_{s},r_{p},e_{o},t)\}$, where $e_{s}, e_{o}\in\mathcal{E}$ and $r_{p}\in\mathcal{R}$. A fact $\{(e_{s},r_{p},e_{o},t)\}$ represents that relation $r$ exists between subject node $e_s$ and object node $e_{o}$ at time $t$. $t$ is sometimes unknown and can be represented in different forms. $r$ can be an entity-entity relation $r_{en\mbox{-}en}\in\mathcal{R}_{en\mbox{-}en}$, or an event-event relation $r_{ev\mbox{-}ev}\in\mathcal{R}_{ev\mbox{-}ev}$, or a relation $r_{en\mbox{-}ev}\in\mathcal{R}_{en\mbox{-}ev}$ between an entity and an event, where $\mathcal{R}=\mathcal{R}_{en\mbox{-}en}\cup\mathcal{R}_{ev\mbox{-}ev}\cup\mathcal{R}_{en\mbox{-}ev}$.
}

EEG~\cite{eeg} is an EKG, where event nodes are abstract, generalized, and semantically complete verb phrases and edges represent temporal and causal relationships between them.  EventKG~\cite{eventkg} is a multilingual event-centric TKG that aggregates event-centric information and temporal relations for historical and contemporary events. ELG~\cite{elg} is an EKG, which mainly focuses on schema-level event knowledge. Besides, EvGraph~\cite{event-centric} is an automatical construction pipeline that structures news stories into EKGs. Compared with TKGs, EKGs contain one more node type (i.e., event nodes) and two more relation types (entity-event relations, event-event relations). Thus, \textbf{Definition 4} subsumes \textbf{Definition 3}.

Basic applications on EKG include script event prediction, timeline generation, and inductive reasoning~\cite{eventkg_review}. Furthermore, EKG can facilitate downstream event-related applications such as financial quantitative investments and text generation.

\section{Knowledge Extraction}\label{sec:3}
The construction of different types of KGs relies on various knowledge extraction techniques. The construction of SKGs requires factual triples extracted from text. Therefore, their knowledge extraction processes mainly include named entity recognition (NER) and relation extraction (RE). Meanwhile, the construction of DKGs requires dynamic knowledge extraction from the ever-changing real-world information.
The construction of TKGs and EKGs additionally requires event extraction to obtain the corresponding temporal and event information. Besides, logical rules between relations in KGs can be regarded as a special kind of knowledge in addition to factual knowledge and rule discovery over different types of KGs requires different rule extraction techniques. With the rise of large language models (LLMs) and their strong natural language understanding capabilities, LLMs are widely used in various knowledge extraction tasks. In this section, we will review the above-mentioned knowledge extraction techniques and the applications of LLMs in these techniques respectively.

\subsection{Static Knowledge Extraction}\label{sec:3.1}
\subsubsection{Named Entity Recognition}\label{sec:3.1.1}
Named Entity Recognition (NER) is a crucial task in text processing, identifying entities like dates, persons, and organizations. Traditional sequence tagging methods like CRF~\cite{CRF} and Pointer Network~\cite{PointerNetwork} have evolved with the advent of deep learning, leading to Bi-LSTM and CNN-based models. These DNN models, like the ones proposed by Andrej et al.\cite{BiLSTM_NER} and Emma et al.\cite{CNN_based_NER}, show significant improvements in NER tasks.

The emergence of language models (LM) brought new dimensions to NER, with discriminative models for word prediction and classification, and generative models for sequence generation. LMs are divided into static and contextualized embeddings, with the former represented by GloVe~\cite{glove} and word2vec~\cite{wordembed1}, and the latter by advanced models like Flair~\cite{flair}, ELMo~\cite{ELMO}, and BERT~\cite{bert}.

Word2vec~\cite{wordembed1}, GloVe~\cite{glove}, and fastText~\cite{fasttext1} represent early milestones in word representation, providing the foundation for subsequent NER models. These embeddings, used as inputs to decoding systems, have been proven effective in various NER tasks, as demonstrated by Nguyen et al.\cite{word2vec_NER} and Ma et al.\cite{glove_NER}. However, their static nature limits contextual adaptability, leading to the development of contextualized LM embeddings.

Contextualized embeddings, like Flair~\cite{flair} for character-level and ELMo~\cite{ELMO} and BERT~\cite{bert} for word-level analysis, have enhanced NER performance by incorporating context. Character-level LMs like Flair~\cite{flair} and word-level LMs like ELMo and BERT have shown remarkable effectiveness in various NER datasets, emphasizing the importance of context in entity recognition.

Generative LMs like GPT~\cite{GPT} and BART~\cite{BART} have redefined NER as a sequence-to-sequence task. These models generate output sequences with identified entities and their types, as explored by Ben et al.\cite{ANL_NER} and Paolini et al.\cite{TANL}. While these methods leverage the generative power of LMs, ensuring precision in entity generation remains a challenge.


\subsubsection{Relation Extraction}\label{sec:3.1.2}
Relation Extraction (RE) is key in identifying connections between mentioned entities in text.
It's broadly categorized into pipeline and joint extraction methods.
The pipeline approach sequentially recognizes entities and then extracts their relations, treating the tasks independently.
In contrast, the joint approach combines NER and RE, aiming to produce comprehensive entity-relation triples.

Pipeline methods in RE include feature-based and kernel-based approaches, treating RE as a multi-class text classification task. Rink et al.\cite{RE_featurebased} use lexical and semantic features for relation classification, achieving significant accuracy. Tazvan et al.\cite{RE_kernelbased} apply a kernel-based approach with subsequence patterns, showing effectiveness in two corpora. Deep learning has further enhanced RE, minimizing the need for feature engineering. Zeng et al.\cite{RE_NNbased} employ convolutional neural networks, outperforming traditional methods. Distant supervision, like the PCNNs model\cite{distant}, addresses data scarcity. Language model-driven RE has recently gained attention. Wu et al.\cite{RE_bert} adapts BERT for RE, and Wu et al.\cite{RE_TRE} further develop TRE, a self-attentive architecture model. Soares et al.\cite{RE_match} proposes task-agnostic relation representations, and generative models like REBEL \cite{REBEL} and copyRLL~\cite{copyRLL} show promising results.

In joint extraction, deep learning (non-LM) and LM-based methods are prevalent. Miwa et al.\cite{RE_LSTM_RNN} present an LSTM-RNN model for joint extraction, achieving improvements on ACE datasets. Katiyar et al.\cite{RE_joint_LSTM} introduce an LSTM with attention, outperforming the SPTree model. Yu et al.\cite{RE_Joint_NDS} propose a sequence labeling approach for joint extraction, demonstrating state-of-the-art performance. LM-based joint extraction employs pre-trained models for encoding. Wei et al.\cite{RE_joint_CASREL} use BERT-based encoding, surpassing baselines on NYT and WebNLG datasets. Eberts et al.\cite{SpERT}introduce SpERT, a BERT-embedded attention model, showing pre-training benefits. Sui et al.\cite{RE_joint_SetPredictionNetworks} use BERT for sentence encoding, achieving impressive results. Generative models like CGT \cite{CGT} and TEMPGEN\cite{TEMPGEN} enhance joint extraction with novel techniques, showing enhanced outcomes.


\subsection{Dynamic Knowledge Extraction}\label{sec:sec3.5}

\begin{figure}[t]
	\centering
	\includegraphics[width=1\columnwidth]{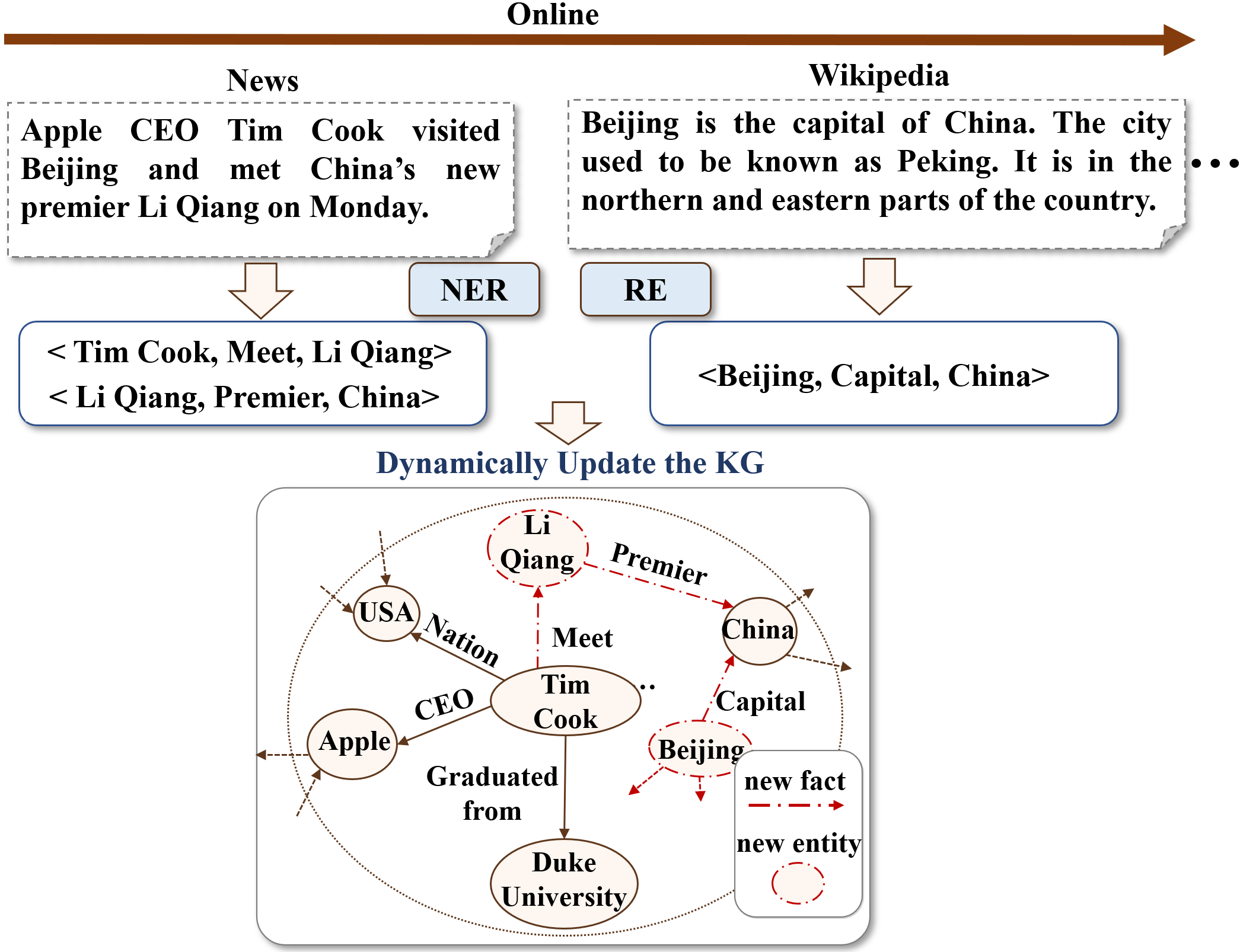}
	\caption{An example of dynamic knowledge extraction. By NER and RE, triples are extracted from all kinds of open information, and the knowledge graph is dynamically updated by these new facts and entities.}
	\label{fig:fig_DKE}
\end{figure}

Dynamic knowledge extraction(DKE) extends beyond static datasets to embrace the continually evolving nature of real-world information. This approach is crucial for online, real-time knowledge acquisition from open sources, as illustrated in Figure~\ref{fig:fig_DKE}. Dynamic knowledge extraction builds upon static KE methods like NER and RE techniques, posing new challenges and requirements in handling the fluidity of information. 
 We introduce two primary methods in DKE: automatic extraction and never-ending learning.

In automatic extraction, systems like KnowItAll by Etzioni et al.\cite{knowitall} autonomously harvest web facts, accumulating over 54,753 facts in four days. Ollie by Schemitz et al.\cite{ollie} addresses limitations in relation extraction, demonstrating improved AUC scores. "Knowledge Vault" by Xin et al.~\cite{knowledge-vault} combines web-extracted information with existing knowledge bases to generate dynamic knowledge graphs.

In never-ending learning, Carlson et al.\cite{Carlson_NELL} propose the "Never-Ending Language Learner" (NELL), continuously extracting web information and refining its learning process. Mitchell et al.\cite{Never_ending_learning} introduce a system for perpetual information acquisition and reading comprehension enhancement. NELL, active since 2010, has amassed over 80 million facts by 2015, with its growth trackable via URLs and Twitter.


\subsection{Temporal Knowledge Extraction}\label{sec:temporalKE}
Temporal Knowledge Extraction (TKE) focuses on understanding and extracting  time-related information from text.  TKE not only identifies the temporal entities but also understands their relationships and contextual implications. The ultimate goal of TKE is to enrich KGs with temporal dimensions. 

The first step in TKE is recognizing temporal entities, such as dates, times, and durations. An example of early work in this domain is TimeML~\cite{pustejovsky2003timeml}, a rich specification language for temporal annotation. Following this, more advanced neural models have been developed. For instance, Chambers et al.\cite{chambers2013event} proposed a method to extract temporal orderings of events using unsupervised learning. Further advancements led to the integration of deep learning techniques. 

Beyond recognizing temporal entities, TKE involves understanding their interrelations. A significant contribution in this area is by Ning et al.\cite{ning2018cogcomptime}, who presented a cognitive computation approach to infer temporal relations. Their work showcased the use of structured learning algorithms to predict temporal links between events. More recently, the focus has shifted towards end-to-end models. For instance, Vashishtha et al.\cite{vashishtha2020temporal} developed a neural network model that can simultaneously perform entity recognition and relation extraction.


\subsection{Event Knowledge Extraction}\label{sec:sec3.4}

\begin{figure}[t]
	\centering
	\includegraphics[width=0.6\columnwidth]{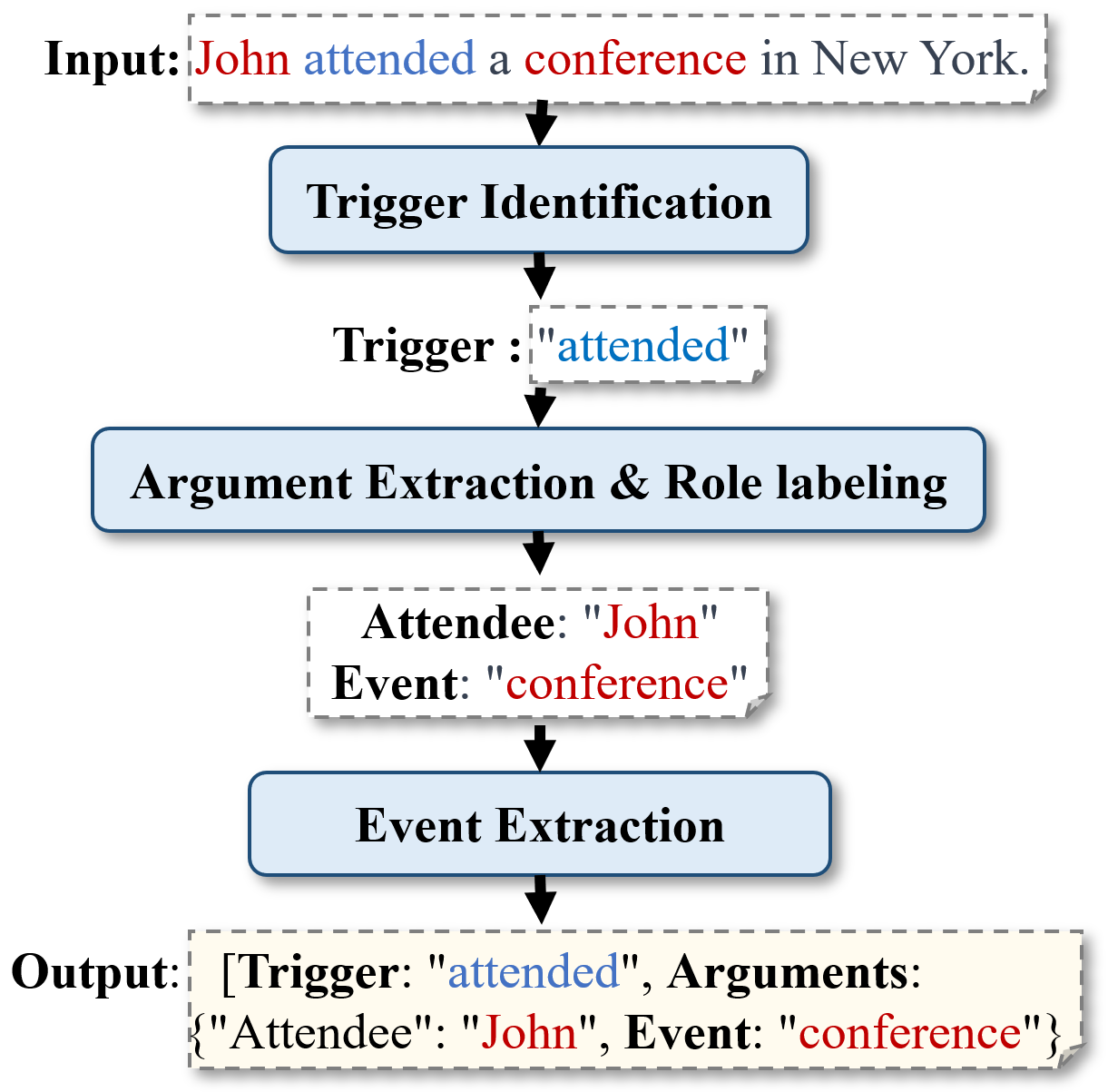}
	\caption{Examples of Event Extraction.}
	\label{fig:fig_ee}
\end{figure}



Event extraction involves identifying events and related information from texts, requiring the detection of triggers, identification of event arguments, and their coherent linkage. This task spans from open-domain to closed-domain extraction, with trigger detection and argument identification as core components. Techniques like rule-based methods, machine learning, syntactic parsing, and semantic role labeling are commonly used. Enhancements include labeled data training, synthetic data generation, graph-based models, and causal inference for improved accuracy in complex event relationships.

Traditional event extraction methods rely on handcrafted features for identifying event triggers and arguments, using machine learning models like RNNs and CNNs for classification. Nguyen et al.\cite{nguyen2016modeling} employs an RNN for trigger detection, while Ghaeini et al.\cite{ghaeini2018event} uses CNNs with skip-gram models for context capture. Nguyen et al.\cite{nguyen2018graph} introduces GCNs for capturing event trigger-argument dependencies. Recent advancements include the Conditional Markov Neural Logic Network by Du et al.\cite{du2021excar}, integrating causal relationships and knowledge graphs for event analysis.

The advent of large language models like BERT and GPT has significantly enhanced event extraction. Yang et al.\cite{yang2019exploring} demonstrates improvements in event extraction using BERT. Tang et al.\cite{tang2020exploring} proposes a multitask learning approach for event classification and explanation. Chen et al.\cite{chen2022ergo} introduces ERGO, a graph transformer for event causality, while Li et al.\cite{li2022kipt} presented KiPT, using knowledge-injected prompts. Du et al.\cite{du2022graph} and Zhou et al.\cite{zhou2022eventbert} leverages graph-enhanced models and pre-trained models like EventBERT for event correlation reasoning. Cross-lingual extraction and continual pre-training for temporal reasoning are also explored, as in Fincke et al.\cite{fincke2022language} and Han et al.\cite{han2020econet}.

\begin{table*}[t]
\begin{center}
\caption{Overview of Knowledge extraction methods}\label{table:KG_extraction}
\footnotesize
\renewcommand{\arraystretch}{0.8} 
\setlength{\tabcolsep}{4pt} 
\begin{tabular}{p{2.5cm}|l|p{2.5cm}|p{6.5cm}|p{4cm}}
\toprule
 \textbf{Extraction Techniques}& \textbf{KG Type} & \textbf{Method Category} & \textbf{Description} & \textbf{Representative Methods} \cr 
 
\midrule
 \multirow{3}{=}{Named Entity Recognition}&\multirow{3}{*}{All types}& Static \& Discriminative&Static embeddings are utilized in sequence labeling tasks and taggers are employed to identify and consolidate entities& word2vec \cite{word2vec_NER}, GloVe\cite{glove_NER},\cite{glove_NER2}, fastText\cite{fasttext_NER}... \cr
 \cmidrule{3-5}
 & & Contextualized \& Discriminative & Contextualized embeddings capture the contextual information in a sequence & Flair\cite{flair}, BERT\cite{bert_NER}, ELMo\cite{elmo_ner} ... \cr
 \cmidrule{3-5}
 & & Contextualized \& Generative & Additional token markers are generated to represent the entity types within the generative sequence & ANL\cite{ANL_NER}, TANL\cite{TANL}, DEEPSTRUCT\cite{deepstruct}, UGF\cite{UGF_NER}... \cr
 
\midrule
\multirow{2}{=}{Relation Extraction}&\multirow{2}{*}{All types}& Pipeline-based& As a pipeline, entities are initially identified, followed by the subsequent relation classification & Feature-based\cite{RE_featurebased}, Kernel-based\cite{RE_kernelbased}, CNN-based\cite{RE_NNbased}, PCNN+MIL\cite{distant}, BERT\cite{RE_bert}, TRE\cite{RE_TRE}, Soares, et al.\cite{RE_match}, REBEL\cite{REBEL}, CopyRLL\cite{copyRLL}... \cr
\cmidrule{3-5}
 & & Joint-based & Triples comprising entities and their corresponding relations are jointly generated as output & SPTree\cite{RE_LSTM_RNN}, Katiyar et al.\cite{RE_joint_LSTM}, NDS\cite{RE_Joint_NDS}, CASREL\cite{RE_joint_CASREL}, SpERT\cite{SpERT}, Sui et al.\cite{RE_joint_SetPredictionNetworks}, CGT\cite{CGT}, TEMPGEN\cite{TEMPGEN}...
 \cr

\midrule
\multirow{1}{=}{Dynamic Knowledge Extraction}&\multirow{1}{*}{DKG}& --- &Entities are relations are dynamically or automatically extracted from open information & KnownItAll\cite{knowitall}, Ollie\cite{ollie}, Knowledge Vault\cite{knowledge-vault}, NELL\cite{Carlson_NELL}, \cite{Never_ending_learning}...\cr

\midrule
\multirow{1}{=}{Temporal Knowledge Extraction}&\multirow{1}{*}{TKG}& --- &Entities are relations contains temporal information & TimeML~\cite{pustejovsky2003timeml}, Chambers et al.\cite{chambers2013event}, Ning et al.\cite{ning2018cogcomptime}, Vashishtha et al.\cite{vashishtha2020temporal}...\cr

\midrule
 \multirow{2}{=}{Event Extraction}&\multirow{3}{*}{EKG}& Neural network based&Neural networks such as RNN classify text utterances at the token level. For document-level event extraction it performs better & ED-CNN\cite{nguyen2016modeling}, GCN-ED\cite{ghaeini2018event}, FBRNNs\cite{chen2016event} \cr
 \cmidrule{3-5}
 & & Large Language Model-based &Large Language Model conducts multi-stage extraction of a sentence by a powerful model transfer ability, For low resource case event extraction it performs better. & EventBERT\cite{zhou2022eventbert}, PLMEE\cite{yang2019exploring} \cr


\bottomrule
\end{tabular}
\end{center}
\end{table*}

\section{Knowledge Reasoning}\label{sec:4}
Considering the intrinsic limitations of collected data, knowledge extraction alone is insufficient to capture the full spectrum of knowledge. These limitations underscore the need for KG reasoning to supplement the KG by inferring missing entities and relationships. Moreover, KGR also enhances downstream tasks by leveraging the constructed KG. Thus, KGR is also essential for creating a comprehensive and effective KG.

Early work on KGR mainly focus on SKGR. However, these approaches are not applicable to other types of KGs. To this end, multiple works are subsequently proposed for DKGR and TKGR. In this chapter, we review all three types of KGR techniques.

\subsection{Static Knowledge Graph Reasoning}\label{sec:4.1}
In this section, we present an overview of classic papers on static knowledge inference in knowledge graphs. These approaches can be categorized into five primary groups: matrix decomposition-based, translation-based, graph neural network-based, large language model-based, and neural-symbolic based methods.
The formal definition is declared as follows,
\par\smallskip

\noindent\textbf{Definition 5. Static Knowledge Graph Reasoning.}
~\textit{
Given a static knowledge graph $\mathcal{SKG}=(\mathcal{E}{n},\mathcal{R},\mathcal{F})$, Static Knowledge Graph Reasoning is a task that seeks to utilize the existing facts encapsulated in $\mathcal{F}$ to infer a queried fact $(e_s^q, r_p^q, e_o^q)$, where $e_s^q, e_o^q\in\mathcal{E}_n$ and $r_p^q\in\mathcal{R}$.
}

\subsubsection{Matrix Decomposition-based Method}\label{sec:4.1.1}
Matrix decomposition-based techniques have been widely employed for static knowledge reasoning in knowledge graphs. By decomposing the adjacency matrix or other related matrices of the knowledge graph into latent factors, these methods represent entities and relations in a low-dimensional latent space, enabling efficient computation of similarity and relationship prediction.

RESCAL~\cite{RESCAL} is a classical matrix decomposition-based knowledge reasoning method, which employs a three-way tensor factorization method for modeling multi-relational data in knowledge graphs. By capturing the interactions between latent factors for each entity and relation, RESCAL has demonstrated superior performance in knowledge reasoning. It can be formally expressed as follows:

$$
\hat{Y}_{ijk} = \mathbf{e}_i^T \mathbf{R}_k \mathbf{e}_j,
$$
where $\hat{Y}_{ijk}$ signifies the predicted score for the triple $(i, j, k)$, with $i$ and $j$ representing entities and k indicating a relation. $\mathbf{e}_i$ and $\mathbf{e}_j$ correspond to the embedding vectors of entities i and j, respectively, while $\mathbf{R}_k$ denotes the relation matrix for relation $k$. The superscript $T$ represents the matrix transpose operation.

Considering the high computational complexity of RESCAL, the DisMult~\cite{DISTMULT} was proposed, which is a simplified version of the RESCAL model that represents relations as diagonal matrices. By reducing the number of parameters, DistMult achieves comparable performance to RESCAL but with lower computational complexity.
Considering the differences between entities and relations in static knowledge graphs, the ComplEx~\cite{ComplEx} extends RESCAL by using complex-valued embeddings to represent entities and relations. By capturing asymmetric relationships through complex conjugate operations, ComplEx can effectively model knowledge graphs with complex relational structures. 
SimplE~\cite{SimplE} is also a representative matrix decomposition method, which employs a simple bilinear form for modeling relations between entity pairs. SimplE incorporates inverse relations into the model, enabling the learning of symmetric and asymmetric relational patterns in the knowledge graph.

Matrix decomposition-based methods offer advantages in modeling static knowledge reasoning in knowledge graphs by capturing latent semantics and patterns in the data. These approaches enable efficient computation of similarity and relationship prediction through low-dimensional latent space representations. However, they face limitations in terms of scalability and handling sparsity, which may hinder their applicability in large-scale knowledge graphs or those with complex relational structures.

\subsubsection{Translation-based Method}\label{sec:4.1.2}
Translation-based methods represent entities and relationships as vectors in a continuous space, and reasoning is performed by translating entity vectors along relationship vectors.

TransE~\cite{TransE} is a classic and basic algorithm that learns entity and relationship embeddings by minimizing the distance between the head entity plus the relationship and the tail entity for each true triple. Specifically, given the vectors $\mathbf{h}, \mathbf{r}, \mathbf{t}$, the scoring function is:
$$
f(h, r, t)=-\|\mathbf{h}+\mathbf{r}-\mathbf{t}\|_{1 / 2},
$$
where $\|\cdot\|_{1 / 2}$ means that the distance can be $L_1$ or $L_2$ distance.

TransE is simple and computationally efficient, making it a popular choice for knowledge graph reasoning tasks. However, TransE has difficulties modeling complex relationships, such as one-to-many or many-to-many.

Based on the TransE, TransH~\cite{TransH} extends TransE by projecting entities onto relation-specific hyperplanes before performing translations. This allows TransH to model complex relationships more effectively than TransE.
TransR~\cite{TransR} further improves upon TransE and TransH by learning separate entity and relationship spaces. TransR projects entities into relation-specific spaces using relation matrices, allowing it to capture more complex relationship patterns.
RotatE~\cite{RotatE} extends TransE by using complex-valued embeddings and representing relationships as rotations. This allows it to capture more complex relationship patterns while retaining the simplicity and scalability of TransE.

In conclusion, Translation-based methods for knowledge graph reasoning are simple, computationally efficient, and scalable, making them suitable for a wide range of applications. However, they face difficulties in modeling complex relationships and handling noisy or incomplete data.

\subsubsection{GNN-based Method}\label{sec:4.1.3}
This is achieved by gathering and integrating information from neighboring nodes within the knowledge graph. The majority of GNN-based methodologies can be encapsulated by the following formulation:

\begin{equation}
\small
H_t^{l+1} \leftarrow \underset{\forall (s,r,t) \in \mathcal{Q}}{\text{\textbf{AGG}}} \left( \text{\textbf{ATT}}(H_s^{l}, H_r^{l}, H_t^{l}) \cdot \text{\textbf{MSG}}(H_s^{l}) \right) || \text{\textbf{SELF}}(H_t^{l}),
\end{equation}
where $H$ represents the entity embeddings that have been learned. The abbreviations MSG, ATT, AGG, and SELF stand for message passing, attention mechanism, aggregation, and self-loop mechanism, respectively. Each of these components plays a vital role in the successful operation of GNN-based methods.

GCN~\cite{GCN} is the fundamental model of the GNN-based method, however, it lacks of ability to model multi-relational data.
R-GCN~\cite{RGCN} is a pioneering GNN-based method for knowledge graph reasoning. It extends traditional graph convolutional networks to handle multi-relational data. R-GCN effectively captures local graph structure, but it can suffer from scalability issues and over-smoothing in deep architectures.
CompGCN~\cite{CompGCN} explicitly models the composition of relations in knowledge graphs. CompGCN learns embeddings by composing entity and relation embeddings in the neighborhood of each entity. This enables it to better capture complex relationships and reason about compositions of multiple relations.

In conclusion, GNN-based methods for knowledge graph reasoning can effectively capture local graph structure and complex relationships, making them a suitable choice for various applications. However, they may face challenges related to scalability, over-smoothing, computational complexity, and parameter tuning.

\subsubsection{Large language models-based Method}\label{sec:4.1.4}
Large Language Models (LLMs) have exhibited considerable potential in the field of static knowledge graph reasoning. This potential stems from their inherent ability to store and retrieve factual data that is acquired during the pre-training phase. The primary methodology involves converting the triple path into a coherent textual sequence, subsequently utilizing the pre-trained LLMs to carry out reasoning tasks.

BERT~\cite{bert} is one of the widely adopted basic language models for knowledge graph reasoning. 
Based on BERT, K-BERT~\cite{KBERT} is an extension of the BERT model that incorporates external knowledge from knowledge graphs during pre-training. By injecting structured knowledge, K-BERT can perform better on knowledge graph reasoning tasks compared to the original BERT. 

KG-BERT~\cite{KG-BERT} also uses BERT to encode the textual contents of entities and relationships (e.g., their description text and names) to get the contextualized representations of triples. 
Coupled with pre-trained word embedding in LLMs, KG-BERT can easily generalize to unseen components. However, KG-BERT has a large room to improve.
MTL-KGC~\cite{DBLP:conf/coling/KimHKS20} added two fine-tuning tasks based on KG-BERT and made the results better. 
StAR~\cite{StAR} augments the textual encoder BERT with graph embedding techniques to enhance its structured knowledge and solves the problem of the efficiency of KG-BERT by reusing graph elements’ embeddings.
CKBC~\cite{CKBC} encodes the various attention weights into the structural representation of the triples and utilizes the triple classification task to fine-tune BERT, which is the same as KG-BERT. The representations of entities and relations in CKBC are a combination of contextual representations from BERT and structural representations from GNN.
Considering the knowledge required by BERT may vary dynamically based on the particular text, CokeBERT ~\cite{CokeBERT} dynamically selects and embeds contextual knowledge based on the actual context of triples.

To improve the learning efficiency of LLM-based methods, SimKGC~\cite{SimKGC} introduces three types of negatives: in-batch negatives, pre-batch negatives, and self-negatives which act as a simple form of hard negatives. 
SimKGC also proposes a re-ranking strategy to reduce the reliance on semantic match of LLM-based methods.
ERNIE~\cite{ERINE} is another large pre-trained language model that explicitly incorporates knowledge graph information. ERNIE uses a dynamic masking strategy during pre-training to better capture the relationships between entities. This enables ERNIE to perform well on knowledge graph reasoning tasks.

In conclusion, large language models incorporating knowledge graph information show promise for knowledge graph reasoning tasks, capturing rich semantic information and leveraging pre-trained knowledge. However, they may face challenges related to scalability, memory limitations, computational complexity, and prompt design.

\subsubsection{Neural-Symbolic Method}\label{sec:4.1.5}
The above static knowledge graph reasoning method all based on neural network models, which suffer the interpretable and unreliable problems, and can not be adopted in some real-world scenarios such as medical, finance, and military which have high reliability demands.



Recently, there have been studies attempting to combine neural and symbolic ways of knowledge reasoning, aiming to leverage the strengths of both paradigms.
We follow the study~\cite{zhang2021neural} and divide relevant methods into symbolic-driven neural reasoning methods and neural-driven symbolic reasoning methods.

Symbolic-driven neural reasoning methods leverage symbolic reasoning to guide and enhance the learning process of neural networks. 
KALE~\cite{KALE} is a symbolic-driven neural reasoning method that integrates first-order logical rules into neural embeddings. KALE encodes logical rules into the neural model by transforming them into soft constraints, which helps to improve the expressiveness and reasoning capabilities of the embeddings. RUGE~\cite{RUGE} is another symbolic-driven neural reasoning approach that incorporates first-order logical rules into the learning process of neural embeddings. RUGE uses rules to guide the training of embeddings, making them more interpretable and capable of handling complex reasoning tasks. 


Neural-driven symbolic reasoning aims to derive the logic rules, where the neural networks are incorporated to deal with the uncertainty and the ambiguity of data, and also reduce the search space in symbolic reasoning.
PRA~\cite{PRA} is a neural-driven symbolic reasoning method that learns to reason over knowledge graphs by exploring paths between entities. PRA employs a neural network to learn the importance of different paths, allowing it to make predictions about relationships in the graph.
MINERVA~\cite{MINERVA} is a neural-driven symbolic reasoning method that uses a reinforcement learning-based approach to explore a knowledge graph. It employs a neural network to guide the search for relevant paths between entities, enabling it to handle multi-hop reasoning tasks effectively.
GraIL~\cite{GraIL} is another neural-driven symbolic reasoning approach that uses graph neural networks to learn first-order logic rules for knowledge graph reasoning. 

In conclusion, neural-symbolic methods for knowledge graph reasoning offer a more interpretable and explainable way of reasoning, making them suitable for various applications that require transparency and understanding. However, they may face challenges related to scalability, computational complexity, implementation complexity, and sensitivity to data quality.

\subsection{Dynamic Knowledge Graph Reasoning}\label{sec:4.2_}

Traditional knowledge graph reasoning methods are under a closed-world assumption, so these models are designed for a fixed set of entities and relations and are unable to consider unseen components.
However, new entities and relations emerge continuously in the ever-changing real world due to several reasons, such as improvements to the knowledge extraction techniques and refinement/enrichment of information. 
Dynamic knowledge graph reasoning aims to deal with emerging entities and relations, which may influence the semantics of existing entities and relations, and predicts the missing links among emerging entities and existing entities.
The formal definition is declared as follows,
\par\smallskip

\noindent\textbf{Definition 6. Dynamic Knowledge Graph Reasoning.}
~\textit{
Given a dynamic knowledge graph $\mathcal{DKG}=\{\mathcal{E}{n},\mathcal{R},\mathcal{F}\}$, the task of Dynamic Knowledge Graph Reasoning seeks to exploit the existing facts in $\mathcal{F}$, which are updated dynamically, to infer a queried fact $(e_s^q, r_p^q, e_o^q)$, where $e_s^q, e_o^q\in\mathcal{E}_n$ and $r_p^q\in\mathcal{R}$.}

We categorized the existing methods according to their modeling perspectives: GNN-based methods and LLM-based methods. In the GNN-based methods, they use GNN or its variants to model emerging entities and relations and the source of the inductive ability can be further divided into four categories: the structure of graphs, relation rules, embedding aggregators and meta-learning. In the LLM-based methods, the entities and relations of the edges in KGs are usually converted into natural language, serving as the input of LLMs, and the powerful understanding ability of LLMs benefits these models. They can be categorized into two groups: depending on one single edge (triple-level) and depending on paths composed of edges (path-level).

With the changing of DKGs, there is a certain probability that unseen entities and relations are long-tail. Therefore, the representations of unseen relations and entities cannot be sufficiently trained given limited training instances (few-shot setting). Thus, few-shot KGR methods are aimed to improve the generalization ability to infer unseen relations with a few facts.

\subsubsection{GNN-Based Methods}
\begin{figure}[h]
	\centering
	\includegraphics[width=1\columnwidth]{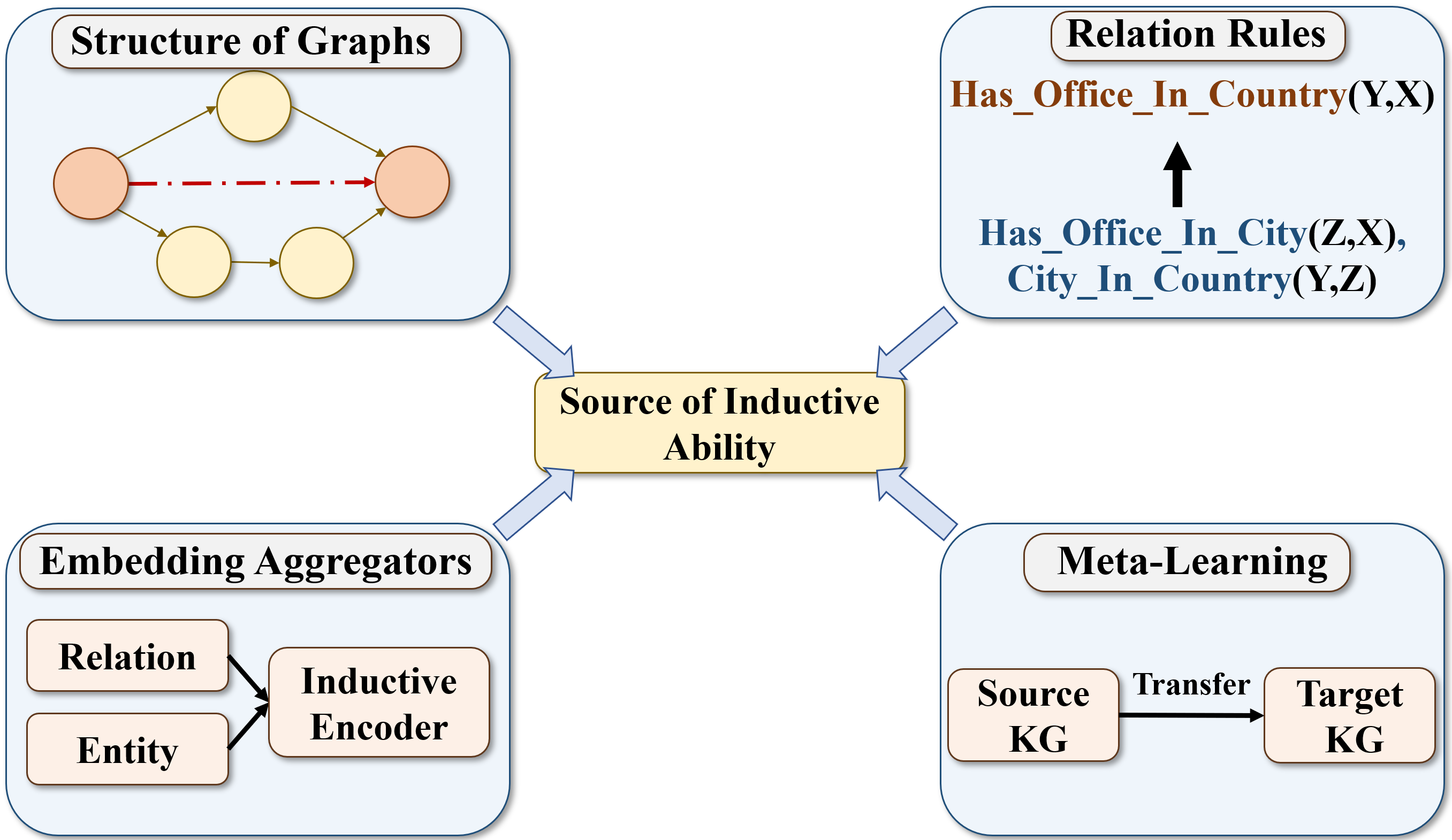}
	\caption{Four sources of inductive ability in GNN-Based DKGR Methods.}
	\label{fig:inductive ability}
\end{figure}

GNN and its variants have been widely used for dynamic knowledge reasoning in dynamic knowledge graphs (DKGs), since GNN can model entities and relations based on the connecting structure and has the potential to handle knowledge out of distribution.
Because of the advantages above, GNN-based DKGR methods capture the transferable knowledge in DKGs through rich structure information and transfer it to new entities and relations, obtaining trusted representations for reasoning tasks over DKGs, such as inductive link prediction. In Figure~\ref{fig:inductive ability}, we categorize the GNN-based methods into four categories, according to four sources of their inductive ability. 

\textbf{Inductive Ability Provided by the Structure of Graphs.}
Subgraphs consisting of the local neighborhood of a particular triplet in DKGs have been proven to contain the logical evidence needed to deduce the relation between the target nodes.
GraIL~\cite{GraIL} proposes to learn entity-independent relational semantics and aggregates the representation of nodes by GNN, so GraIL can be naturally generalized to unseen nodes. GraIL also introduces a series of new fully-inductive benchmark datasets for the inductive relation prediction problem, by sampling disjoint subgraphs from the KGs in four traditional transductive datasets. The subgraph representation $\mathbf{h}_\mathcal{G}{\left(u, v, r_{t}\right)}$ in GraIL is as follows:

$$
\mathbf{h}_\mathcal{G}{\left(u, v, r_{t}\right)}=\frac{1}{|\mathcal{V}|} \sum_{i \in \mathcal{V}} \mathbf{h}_{i},
$$
where $|\mathcal{V}|$ denotes the set of vertices in graph $\mathcal{G}_{\left(u, v, r_{t}\right)}$, $\mathbf{h}_{i}$ is the latent representation of node in subgraph $\mathcal{G}_{\left(u, v, r_{t}\right)}$.
Following GraIL, CoMPILE~\cite{DBLP:conf/aaai/MaiZY021} further considers the directed nature of the extracted subgraph structure to enable a sufficient flow of relation information. Therefore, CoMPILE can naturally handle asymmetric and anti-symmetric relations in subgraphs.
These two approaches require generating one subgraph for each target triple, which is a significant bottleneck in practice.
INDIGO~\cite{DBLP:conf/nips/LiuGHK21} constructs a node-annotated graph, where nodes represent pairs of constants and two nodes are connected only if their pairs share a constant, and the original KG is fully encoded into a GNN in a transparent way. 
Once the types and relations in KGs are fixed, the dimension of nodes in the node-annotated graph does not depend on the input KG, making the inductive capabilities of GNN fully exploited.

\textbf{Inductive Ability Provided by Relation Rules.}
Rule mining uses observed co-occurrences of frequent patterns in KGs to determine logical rules, which can transfer reasoning to unseen facts.

NeuralLP~\cite{yang2017differentiable} is the first end-to-end differentiable approach to learning the structure of logical rules. DRUM~\cite{sadeghian2019drum} further attaches the found rules with confidence scores by bidirectional RNNs.
TACT~\cite{TACT} categorizes topology-aware correlations between relations into seven topological patterns, and learns their importance in an entity-independent manner for inductive link prediction.

PathCon~\cite{DBLP:conf/kdd/0004RL21} leverages relational context and relational paths to directly pass relational messages among edges iteratively to aggregate neighborhood information. PathCon is applicable to the inductive settings since nodes temporarily store the messages from their neighbor edges.
Combining GraIL and PathCon, RMPI~\cite{DBLP:journals/corr/abs-2210-03994} conducts relational message passing over subgraphs for fully-inductive KGC and further proposes new techniques on graph transformation, graph pruning, relation-aware neighborhood attention, addressing empty subgraphs, etc.

\textbf{Inductive Ability Provided by Embedding Aggregators.}
Recent efforts suggest that training a neighborhood aggregator like conventional GNN may help embed new entities inductively via their existing neighbors, so many methods modify GNN as an embedding aggregator in various ways.
LAN~\cite{LAN} introduces a novel aggregator, which addresses the properties by aggregating neighbors with both rules- and network-based attention weights, to facilitate inductive KG embedding.
ROLAND~\cite{DBLP:conf/kdd/YouDL22} views the node embeddings at different GNN layers as hierarchical node states and then recurrently updates them, easily repurposing any static GNN to dynamic graphs.
SE-GNN~\cite{SE-GNN} studies KGE extrapolation problems from a data-relevant and model-independent view and proposes a GNN-based KGE method modeling the semantic evidence.

\textbf{Inductive Ability Provided by Meta-Learning.}
Inspired by the ability of “learning to learn” brought by meta-learning, researchers formulated a set of tasks consisting of support triples and query triples to mimic the link prediction task in the emerging KG, and used GNN to embed unseen components in each task.
HRFN~\cite{10.1145/3459637.3482367} concentrates on effectively embedding out-of-knowledge-base entities by a two-stage model, where coarse-grained pre-representation and fine-tuning aim at emerging entities are used.
MaKEr~\cite{MaKEr} and MorsE~\cite{MorsE} propose a problem called inductive knowledge graph embedding. They train knowledge graph embedding models based on meta-learning on a set of entities that can generalize to unseen entities for either in-KG(i.e., link prediction) or out-of-KG tasks(i.e., question answering). The meta-training objective in MorsE is:
$$
\min _{\theta, \phi, \mathrm{R}} \sum_{i=1}^{m} \mathcal{L}_{i}=\min _{\theta, \phi, \mathrm{R}} \sum_{i=1}^{m} \mathcal{L}\left(Q_{i} \mid f_{\theta, \phi}\left(G_{i}\right), \mathbf{R}\right),
$$
where $\mathrm{R}$ is the relation embedding matrix for all relations, $f_{\theta, \phi}\left(G_{i}\right)$ is used for outputting entity embedding matrix for
entities in the current task, \{ $\theta, \phi$, $\mathrm{R}$\} are learnable parameters;
$m$ denotes the number of all sampled tasks.

In conclusion, GNN-based methods are under an encoder-decoder framework and are able to take advantage of the connectivity structure to generate the representation of emerging entities and relations. GNN and its variants, such as R-GCN and GAT, provide insightful semantic information or rule patterns for these methods and use relation-specific transformation to model the directed nature of DKGs, making GNN-based methods explainability over DGKs, scalability and suitable for various applications. However, GNN-based methods may face challenges related to computational complexity and over-smoothing, and they are restricted by the original framework of GNN.

\subsubsection{LLM-Based Methods}
\begin{figure}[h]
	\centering
	\includegraphics[width=1\columnwidth]{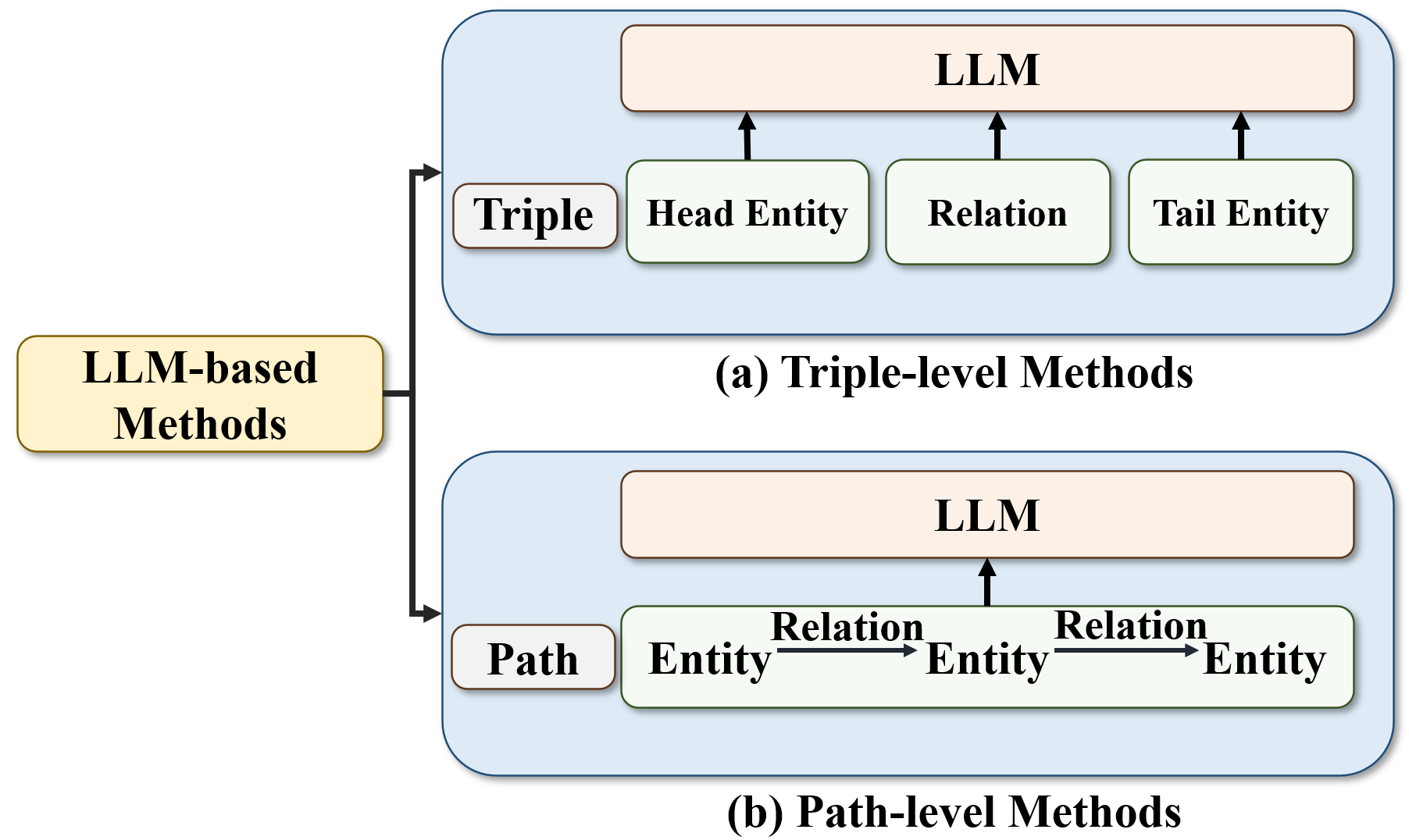}
	\caption{Two categories of LLM-Based DKGR Methods.}
	\label{fig:DKGR_LLM}
\end{figure}

The proposal of Large Language Models (LLMs) indicates that general semantic knowledge which widely exists in open source corpus can be encoded into the parameters of LLMs via pre-training and fine-tuning. Current LLM-based works research how to integrate KG-specific knowledge into the inductive ability of LLMs and attempt to convert from a data-driven way to a knowledge-driven way. In Figure~\ref{fig:DKGR_LLM}, LLM-based DKGR methods are classified into two categories: triple-level methods and path-level methods.

\textbf{Triple-level Methods.}
Before the proposal of LLMs, DKRL~\cite{DBLP:conf/aaai/XieLJLS16} explores the continuous bag-of-words (CBOW) model to maximize the likelihood of predicting the description text of entities. 
LLM-based models mentioned in Section~\ref{sec:4.1.4} like KG-BERT, StAR, CKBC can also be used in DKGR since they are able to deal with unseen entities via the pre-trained prior knowledge in LLMs.
Some recent works conduct experiments in the inductive setting for DKGR.
For example, KEPLER~\cite{KEPLER} encodes textual entity descriptions with BERT as their embeddings and aligns them to the symbol space of KGs, then KEPLER jointly optimizes the knowledge embedding objective and the masked language modeling (MLM) objective. For a a relational triplet $(h, r, t)$, KEPLER encodes each part with BERT E:

$$
\begin{aligned}
\mathbf{h} & =\mathrm{E}_{<\mathrm{s}>}\left(\text { text }_{h}\right) \\
\mathbf{t} & =\mathrm{E}_{<\mathrm{s}>}\left(\text { text }_{t}\right), \\
\mathbf{r} & =\mathbf{T}_{r},
\end{aligned}
$$
where $\text { text }_{h}$ and $\text { text }_{t}$ are the descriptions
for $h$ and $t$, with a special token $<s>$ at the beginning. $T$ is the relation embeddings and
$\mathbf{h}, \mathbf{t}, \mathbf{r}$ are the embeddings for $h, t and r$.
In KEPLER, knowledge embeddings provide factual knowledge for PLMs, while informative text data also benefit knowledge embeddings.
Experimental results illustrate the effectiveness of LLM-based methods in DKGR.

\textbf{Path-level Methods.}
In addition to using BERT for encoding a single triple, there are also many methods leveraging BERT to mine rules in DKGs. The rules here are presented in the form of natural language rather than traditional symbolic rules.

BERTRL~\cite{BERTRL} concatenates the paths consisting of constructed triples into textual information and combines it with the pre-trained prior knowledge, forming rule-like reasoning evidence. The BERT in BERTRL is used as a score function to score the probability that the candidate relational triple is true.
Bi-Link~\cite{Bi-Link} searches for relational prompts according to learned syntactical patterns via BERT and comes up with probabilistic syntax prompts for link predictions. Bi-Link also uses contrastive learning for bidirectional linking between forward prediction and backward prediction to express symmetric relations.

In conclusion, the pre-trained LLMs have shown that factual knowledge inside KGs can be recovered from PLMs, so LLM-based methods convert edges or relation paths connecting subject and object into natural language form. LLMs enable LLM-based DKGs methods to have language comprehension skills and be able to deal with emerging entities and relations in the semantic space. 
However, LLM-based DKGR methods may face the shortcomings of high GPU memory and the static property of LLMs may not catch up with the dynamic property of DKGs for the rapid update of DKGs.

\subsubsection{Few-Shot Knowledge Graph Reasoning}

To learn the representation of the components in KGs, previous methods usually represent relations and entities based on a large amount of training data. In contrast, few-shot knowledge graph reasoning mainly relies on transferring and adapting learned knowledge. In this section, we introduced the original proposal of few-shot learning, a main way of few-shot learning called meta Learning and some other new proposed few-shot learning methods.

\textbf{The Original Proposal of Few-shot Learning.}
GMatching~\cite{DBLP:conf/emnlp/XiongYCGW18} is the first to propose the task of predicting long-tail relations and formulate it as one-shot relational learning.
GMatching learns a matching metric to discover similar triples for target triples. Once trained, GMatching can be adapted to any new relations without fine-tuning, achieving one-shot learning.
For zero-shot(new-added) relations, ZSGAN~\cite{DBLP:conf/aaai/QinWCZXW20} is able to generate text category representations from text descriptions by setting adversarial learning goals.
However, FSRL~\cite{DBLP:conf/aaai/ZhangYHJLC20} thinks works of one-shot and zero-shot learning limit the generalizability of models and do not fully use the supervisory information, so FSRL proposes to discover facts of new relations with few-shot references which is more suitable for practical scenarios. FSRL mainly uses a heterogeneous graph structure encoder based on the attention mechanism and aggregates the few-shot reference set based on LSTM.
FAAN~\cite{FAAN} formally proposes the notion of dynamic properties in few-shot KG completion for the first time and designs an adaptive attention network to adapt dynamic entity representations in different tasks.

\textbf{Meta-learning-based Few-shot Learning Methods.}
Meta-learning for the few-shot setting aims to capture transferable relation-specific meta-information and is suitable and well-performing for dynamic knowledge graphs, since emerging relations usually have few links.
MetaR~\cite{MetaR} proposes relation meta and gradient meta with common and shares knowledge to predict new facts about the target relation, which only has a few observed associative triples.
The gradient meta $G_{\mathcal{T}_{r}}$ for relation meta $R_{\mathcal{T}_{r}}$ in MetaR is regarded as the gradient of $R_{\mathcal{T}_{r}}$ based on the loss $L$, so that relation meta can be updated as follows:

$$
\begin{aligned}
G_{\mathcal{T}_{r}}=\nabla_{R_{\mathcal{T}_{r}}} L\\
R_{\mathcal{T}_{r}}^{\prime}=R_{\mathcal{T}_{r}}-\beta G_{\mathcal{T}_{r}},
\end{aligned}
$$
where $\beta$ indicates the step size of the gradient meta when operating on relation meta.

Meta-KGR~\cite{Meta-KGR} proposes a meta-based RL framework, where the learned meta-knowledge on normal relations provides well-initialized parameters of new models for few-shot relations and the multi-hop reasoning offers interpretability for the results. 
Based on Meta-KGR, FIRE~\cite{DBLP:conf/emnlp/Zhang0S0C20} further performs heterogeneous structure encoding and knowledge-aware search space pruning in the reinforcement learning process, outperforming Meta-KGR.

\textbf{New Proposed Few-shot Learning Methods.}
GEN~\cite{GEN} proposes an inductive learning module to extrapolate knowledge in the support set and then considers inter-relationships between unseen entities in the transductive learning module.
P-INT~\cite{DBLP:conf/emnlp/XuZKDCLL21} leverages the paths that can expressively encode the relation of two few-shot entities and calculates the interactions of the paths.
ADK-KG~\cite{DBLP:conf/sdm/0002QYZ22} uses pre-trained entity and relation embeddings and the optimized parameters can be fast adapted for few-shot relations.
Niu et al.~\cite{DBLP:conf/sigir/NiuLTGDLWSHS21} proposes a gated neighbor aggregator to filter the noise neighbors, capturing the most valuable contextual semantics of few-shot relations, and considers complex relations simultaneously.
REFORM~\cite{REFORM} points out that plenty of errors could be inevitably incorporated in most existing large-scale KGs and proposes an error mitigation module to study error-aware few-shot KG completion.
GMUC~\cite{DBLP:conf/dasfaa/ZhangWQ21} leverages Gaussian metric learning to capture uncertain semantic information and to complete few-shot uncertain knowledge graphs.
Yao et al.~\cite{DBLP:conf/coling/Yao0X022} proposes a data augmentation technique from inter-task view to generate new tasks for few-shot knowledge graph completion, and from intra-task view to enrich the support or query set for an individual task. The proposed framework can be applied to a number of existing few-shot knowledge graph completion models.

In conclusion, few-shot DKG reasoning is a relatively new research area and the meta-learning approach fixes the challenges of data limitation, to a certain extent. However, the dynamic properties in few-shot learning have not been widely researched. Therefore, there is much room to exploit for researchers in the future.

\subsection{Temporal Knowledge Graph Reasoning}\label{sec:4.3}
Both static or dynamic knowledge graphs only record the knowledge graph at a certain time or the latest state, and cannot reflect the evolution process of the relations between entities at different times. Different from them, the temporal knowledge graph records the valid time of knowledge. 
 The formal definition of temporal knowledge graph reasoning (TKGR) is defined as follows:
\par\smallskip

\noindent\textbf{Definition 7. Temporal Knowledge Graph Reasoning.}
~\textit{
Given a temporal knowledge graph $\mathcal{TKG}=\{\mathcal{E}_{n},\mathcal{R},\mathcal{T},\mathcal{F}\}$, the task of Temporal Knowledge Graph Reasoning aims to utilize the existing facts in $\mathcal{F}$, each with associated temporal information from $\mathcal{T}$, to infer a queried fact $(e_s^q, r_p^q, e_o^q, t^q)$, where $e_s^q, e_o^q\in \mathcal{E}_{n}$, $r_p^q\in\mathcal{R}$ and $t^q\in\mathcal{T}$.}

\begin{figure}[h]
	\centering
	\includegraphics[width=1\columnwidth]{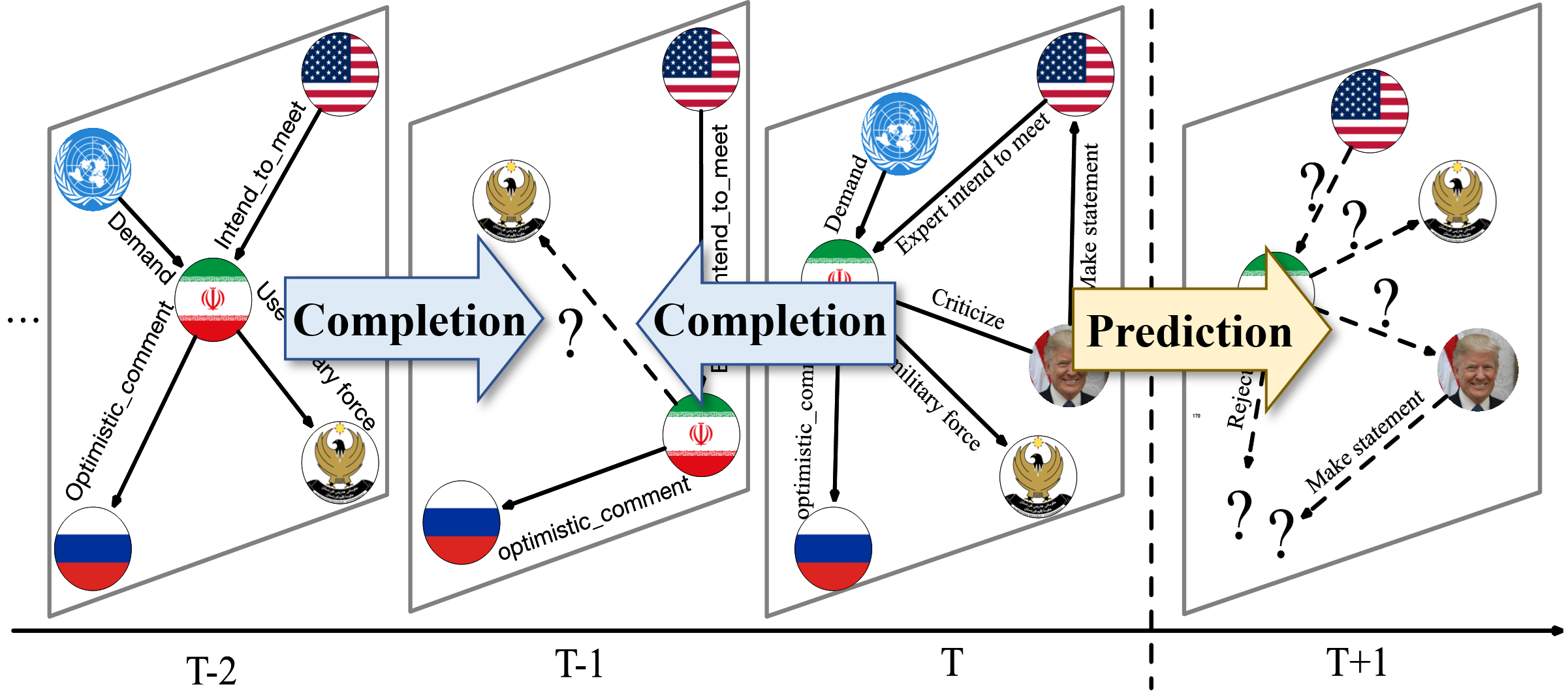}
	\caption{The illustration of temporal knowledge graph completion and temporal knowledge graph prediction.}
	\label{fig:tkgc_tkgp}
\end{figure}

The temporal knowledge graph reasoning includes two parts as shown in Figure~\ref{fig:tkgc_tkgp}: The temporal knowledge graph completion(explaining the TKG at some past time), and the temporal knowledge graph prediction (predicting the state of TKG at a given future time)~\cite{vila1994survey}.

In this section, we classify the methods for temporal knowledge graph reasoning into two categories: completion methods and prediction methods.


\subsubsection{Temporal Knowledge Graph Completion}
The temporal knowledge graph completion methods aim to understand the time information and integrate it into the semantic and structural information in the knowledge graph. The temporal knowledge graph completion methods can be divided into two parts: tensor decomposition-based completion methods and translation-based completion methods.

\textbf{Tensor Decomposition-based Completion Methods.}
The tensor decomposition-based completion methods learn the evolution law of the entire graph. As the temporal knowledge graph is represented as a sequence of static snapshots, which can be modeled as a sequence of matrices or high-dimensional tensors. The tensor decomposition-based completion methods can be divided into CP decomposition methods~\cite{1927The} and Tucker decomposition methods~\cite{tucker1966some}. The former decomposes high-dimensional tensors into a series of rank tensors, and the latter considers the core tensor, which represents the main properties of the original tensor. When the core tensor is a hyper-diagonal matrix, Tucker decomposition is a special case of CP decomposition.

Lacroix et al.~\cite{lacroix2020tensor} represented the temporal knowledge graph as a four-dimensional tensor 
and extends the ComplEX~\cite{ComplEx} to the temporal knowledge graph reasoning, proposing the CP decomposition method TComplEX. Compared to CP decomposition, Tucker decomposition has better reasoning performance, but its decomposition efficiency is lower due to the operations involving the core tensor.  ~\cite{tresp2017embedding} and ConT~\cite{ma2019embedding} used a 4th-order Tucker decomposition to obtain the representations of entities, relations, and time information. ~\cite{shao2022tucker} adds temporal regularization constraints in adjacent time representation based on Tucker decomposition. Considering that different relations may have different temporal properties, such as the periodic feature of relations, and temporal dependencies among different relations. Based on   ~\cite{tresp2017embedding}, TIMEPLEX~\cite{TIMEPLEX} considers the periodicity scores and temporal dependency scores of the relations in the reasoning procedure.


\textbf{Translation-based Completion Methods.}
The representations of entities, relations, and temporal information are not decomposed from snapshots of the knowledge graph but are learned in the form of vectors or other representations.
Translation-based reasoning methods can be divided into two types: reasoning with explicitly represented time information and implicitly represented time information.

\textit{Completion with Explicitly Represented Time Information.}
Reasoning methods with explicit time information represent the temporal information in TKGs in the form of temporal transition matrices, embedded vectors, and time mapping matrices.
Time representation is independent of the representation of entities and relations and represents the evolutionary information of the TKG at each snapshot from a macroscopic level.
The representation of entities and relations is static and remains unchanged, independent of time information.

\cite{TAE2016} observes that the temporal knowledge sequence of the same entity: \textit{(Albert Einstein, born in, Ulm, 1879)}, \textit{(Albert Einstein, awarded, Nobel Prize in Physics, 1922)}, and \textit{(Albert Einstein, passed away in, Princeton, 1955)} represents the temporal sequence of \textit{born in, awarded, passed away}. The time-aware embedding (TAE) incorporates the temporal information of relationships using a temporal matrix to learn relationship representations that contain temporal information.  TAE-ILP~\cite{jiang2016towards} considers the valid time intervals of knowledge and uses the Integer Linear Programming technique to restrict the temporal knowledge of shared head entities to non-overlapping time intervals.
TTransE ~\cite{leblay} learns the time representation based on the static knowledge reasoning method TransE, and combines the representation of entities, relations, and time information to calculate the confidence of the temporal knowledge. 

\begin{figure}[h]
	\centering
	\includegraphics[width=1\columnwidth]{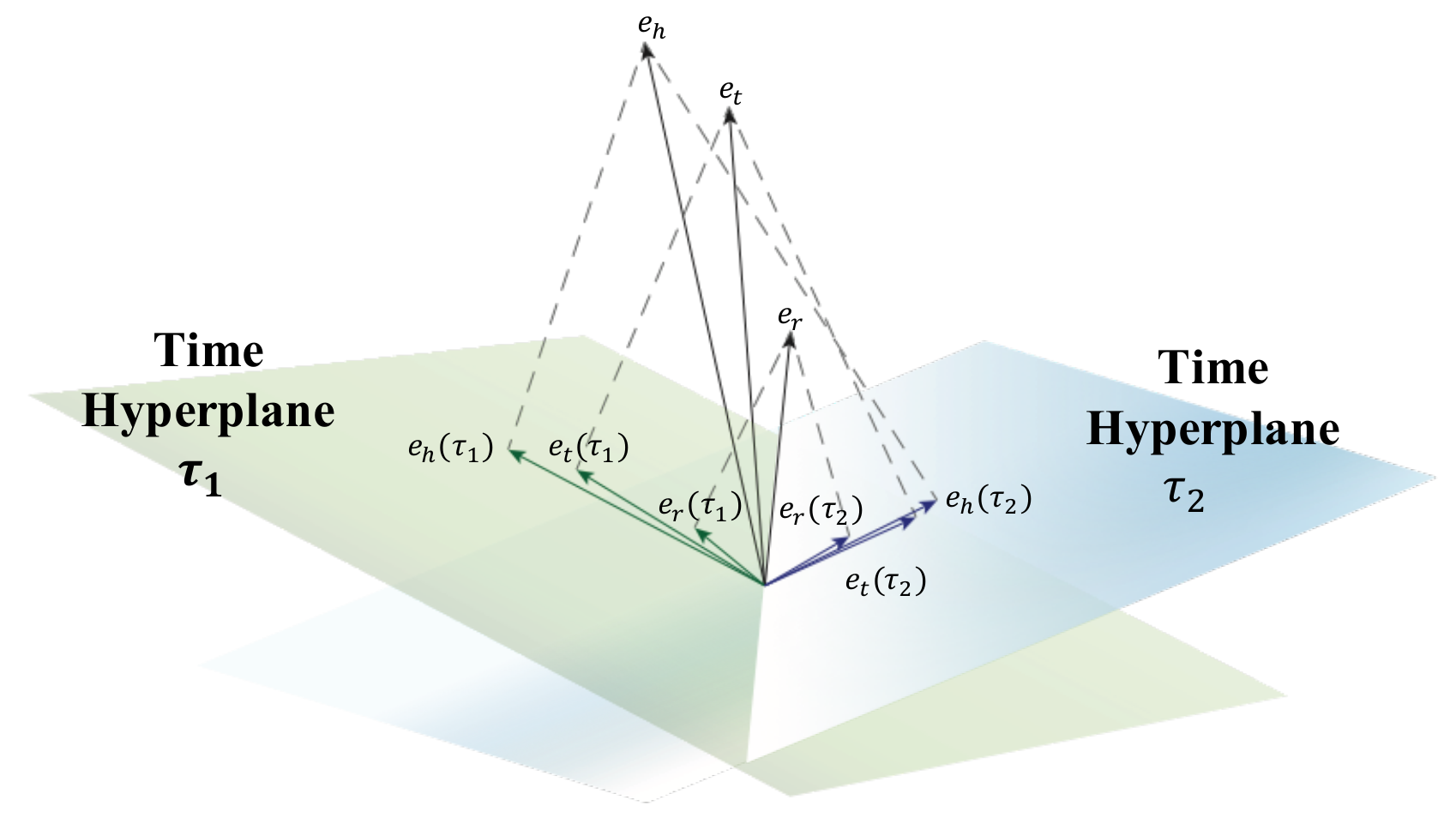}
	\caption{The illustration of HyTE.}
	\label{fig:figure_HyTE}
\end{figure}

HyTE \cite{HyTE} designs the time hyperplane to reflect the entity and relation information at a specific time. As shown in Figure~\ref{fig:figure_HyTE}, in the reasoning process, HyTE projects the head entity, relationship, and tail entity vectors $e_h$, $e_r$, and $e_t$ to the corresponding time hyperplane $T_1$, $T_2$ to obtain the entity and relationship representation at different time.
To reflect the evolutionary process of entity information in the temporal knowledge graph, TeRo~\cite{TERO} defines the entity representation at different moments as rotation in the complex vector space, which is the Hermitian dot product of entity and time vectors, to reflect the evolutionary process of entities.


Considering time information in knowledge graphs can be represented at different granularities, TOKEI~\cite{leblay2020towards} designs the hierarchical temporal matrix that is adaptable to different time granularities  (e.g., year-month-day or year-quarter-month-week-day).
TA-TRANSE~\cite{TA-TransE}  takes into account that time sequences at different granularities are of unequal lengths, such as [\textit{1989}] and [\textit{1989, January, 16th}]. It uses the LSTM to encode the time sequences as the temporal information of knowledge.
However, the aforementioned methods ignore the correlation between temporal information. TeLM~\cite{TeLM}  introduces a temporal smoothing mechanism into the loss function, which constrains the change in the representation of adjacent moments to reflect the temporal correlation.  TIME2BOX~\cite{cai2021time} and RTGE~\cite{RTGE} both design temporal smoothing mechanisms to constrain the change in adjacent time vectors. 


\textit{Completion with Implicitly Time Information.}
The completion methods implicitly represented the time information, and model the evolution of entities or relations with time as functions, which reflects the evolution trend of each entity at each moment. The time information is incorporated in the representation of entities or relationships. 

TRESCAL~\cite{leblay} extends the static knowledge graph reasoning method RESCAL~\cite{RESCAL} to temporal knowledge graph reasoning. TRESCAL considers that the information of relations evolves over time, and the temporal information depends on the existence of the relationship, i.e., the relation-temporal matrix $W_{r,t}$ is trained at different temporal information. 
Diachronic embedding ~\cite{DE-SimplE} regards representations of entity and relation as a series of time functions, which use parameterized Fourier functions to encode the entity and relation representations at time spots.

Furthermore, ATiSE~\cite{ATiSE} considers uncertainty in the time functions and adds random noise to the representation of entities and relations to simulate random interference factors. It conducts reasoning by learning the parameters of a multi-dimensional Gaussian distribution composed of three factors: linear trend, periodic trend, and noise. 
T-GAP~\cite{T-GAP} designs an interpretable mechanism: starting from the target entity $u$ in the incomplete knowledge $(u,r,?,t)$, T-GAP iteratively expands the subgraph using attention pruning strategies and updates the representations and attention scores of each entity in the subgraph through a time-aware graph neural network. 
After the expansion process ends, the entity with the highest attention score in the subgraph is chosen as the correct answer for completion. 

\subsubsection{Temporal Knowledge Graph Prediction}
Temporal knowledge graph prediction methods aim to learn the evolution patterns of TKG historical information to predict future states.
These methods can be further categorized into three types: path-based prediction methods, point process-based prediction methods, and sequence network-based prediction methods.

\textbf{Path-based Prediction Methods}
Due to the graph nature of temporal knowledge graphs, the sampled time-constrained random walk paths in temporal knowledge graphs can be viewed as multi-hop relations between entities to represent reasoning rules.
The path-based prediction methods of temporal knowledge graphs believe that the temporal walks~\cite{nguyen2018dynamic} in the graph reflect the evolution rules of knowledge.

 In an $L$-length temporal walk  $(v_{1}, v_{2}, t_{1}), (v_{2}, v_{3}, t_{2}), ..., (v_{L}, v_{L+1}, t_L)$, the adjacent temporal edges should satisfy the temporal-ordered constraint $t_1 \leq t_2 \leq ... \leq t_L$, ensuring that the walk contains the temporal dependency information between entities.
TLogic ~\cite{TLogic} considers the temporal properties of multi-hop relation reasoning and proposes the Cyclic Temporal Logical Rule, represented as:

$P_1(u, z_1, t) \wedge P_2(z_1, z_2, t+1) \wedge \ldots \wedge P_n(z_{n-1}, v, t+(n-1)) \\
\rightarrow P(u, v, t+n), $

The Cyclic Temporal Logical Rule (CTL) strictly follows the temporal-ordered constraint to collect temporal walks and represents the evolution feature of multi-hop relation reasoning. TLogic learns the confidence of cyclic temporal logical rules through statistical frequency and selects high-confidence paths as reasoning rules, such as: $(E_1, instigates, E_2, t) \wedge (E_2, declares, E_1, t+1) \wedge (E_1, instigates, E_2, t+2) \rightarrow (E_1, protests, E_2, t+3)$.

To extract reasoning rules from sampled temporal paths, it is necessary to encode temporal information in the paths. 
TITer~\cite{sun2021timetraveler} and CluSTeR~\cite{li2021search} use LSTM-based temporal path encoders to learn temporal order relationships. xERTE~\cite{han2021explainable} collects temporal paths from the target entity and propagates attention scores along the paths while continuously expanding the query subgraph. xERTE designs a temporal relation graph attention mechanism to guide expansion in the direction that benefits query prediction and the expanded subgraph can be viewed as an explanation for predicting the entity.


\textbf{Temporal Point Process-based Prediction Methods.}
The temporal Point Process (TPP) is a probabilistic method for predicting future events. The TPP-based prediction methods treat the occurrences and disappearances of relations between entities as sequences of events, where the event probability depends on the historical evolution characteristics of relevant entities.

In TPP-based prediction methods, the expected existence probability of knowledge at time $t$ is represented as the product of the conditional intensity function and the survival function, i.e., $f(t)=S(t)\lambda(t)$. The survival function $S(t)$ evaluates the conditional probability that events do not exist in the time window $[t, t+\Delta t)$. 
 The conditional intensity function $\lambda(t)$ represents the expected existence probability in the time window $[t, t+\Delta t)$ at time $t$, given the event history information $H(t)$.

$$\lambda (t | H(t)){dt}=\mathbb{E}[N(t, t+dt) | H(t)]$$

Know-evolve~\cite{Know-evolve} first applies TPP to multi-relation reasoning. It treats transient knowledge $(u,r,v,t)$ as a probability event. It uses a loss function based on a multivariate Rayleigh point process to model the event probability.
The Graph Hawkes Neural Network (GHNN)~\cite{han2020graph}  can effectively aggregate information from current entity neighbors to predict the future.

For effectively learning the evolution of temporal knowledge graphs in a data-driven approach rather than the pre-defined conditional intensity function, neural temporal point processes have become a popular research direction for TPP-based prediction methods. 

DyRep~\cite{trivedi2018dyrep} models both long-term(association) and short-term(communication) knowledge between individuals in social networks. The former reflects the changes in the network topology, representing stable relations, while the latter represents temporary, instant interactions. The events that occur are propagated through three information propagation modes: Localized Embedding Propagation, Self-propagation, and Exogenous Drive.
DyRep uses neural networks to model the conditional intensity function $\lambda_k^{(u,v)}(t)$, which calculates the probability of the existence of associated knowledge between entities (u,v) at time t (k denotes the relation type). 



M$^2$DNE~\cite{lu2019temporal} learns entity representations by modeling evolutionary trends at both macro and micro levels. At the macro level, M$^2$DNE imposes temporal dependency constraints on entity representations, such as the number of newly added knowledge at time $t$ is determined by the number of knowledge at $t-1$. At the micro level, M$^2$DNE captures the evolutionary patterns of entity pairs, using a neural network-based conditional intensity function to calculate the probability of knowledge.
Similarly, TDIG~\cite{TDIG} and EvoKG~\cite{EvoKG}  both use graph neural networks and gated recurrent neural networks to model neural temporal point processes for tasks such as relation inference and temporal information reasoning.


\textbf{Sequence Neural Network-based Prediction Methods.}
Due to the powerful learning ability of neural networks, both static and dynamic knowledge reasoning methods based on neural networks have been proven to effectively learn semantic and structural features of knowledge graphs for reasoning. 
Sequence neural network-based prediction methods consist of snapshot encoders and temporal encoders: snapshot encoders extract representations of entities at different times from the snapshot sequence; temporal encoders integrate the snapshot representations to learn the entity evolution information from them.

Based on the types of temporal encoders, these methods can be divided into RNN-based temporal encoding methods and self-attention-based temporal encoding methods.


\textit{RNN-based Temporal Encoding Methods.} RE\_NET~\cite{jin2019recurrent} aggregates the representations of entity neighbors in historical time steps, concatenates them with the vector representations of the head entity $s$ and relation $r$, encodes the evolving information through RNN to obtain the latest representation of the tail entity, which is used to predict the correct entity in the next time step.
To better learn the topological features of knowledge graphs at each stage, RE-GCN~\cite{RE-GCN} uses a relation-aware graph convolutional network (RGCN) to encode snapshot features in the knowledge graph and then learns the evolution patterns of features through a recurrent neural network. 
CEN~\cite{CEN} considers the length diversity and temporal variability of temporal patterns. It encodes multiple lengths using a relation-aware graph convolutional network to model the length diversity and trains the model online to address temporal variability. 
HiSMatch~\cite{HiSMatch} uses relation-aware graph neural networks and recurrent neural networks to distinguish between graph sequences relevant to queries and those relevant to candidate entities and proposes a matching framework for temporal reasoning. 
However, recurrent neural networks suffer from the problem of information forgetting when encoding long-distance temporal information and perform poorly in multi-step prediction tasks~\cite{DGNN5}.


\begin{table*}[t]
\begin{center}
\caption{Overview of Knowledge Graph Reasoning Methods.}\label{table:KG_Reasoning}
\footnotesize
\renewcommand{\arraystretch}{0.8} 
\setlength{\tabcolsep}{4pt} 
\begin{tabular}{l|p{3.6cm}|p{6cm}|p{5cm}}
\toprule
 \textbf{KG Type} & \textbf{Method Category} & \textbf{Description} & \textbf{Representative Methods} \cr 
\midrule
 \multirow{10}{*}{Static KG}& Matrix decomposition-based Methods& Adopt matrix decomposition operations to obtain the embedding of entities and relations & RESCAL\cite{RESCAL}, DISTMULT\cite{DISTMULT}, ComplEx\cite{ComplEx}, SimplE\cite{SimplE}
 \cr
 \cmidrule{2-4}
 & Translation-based Methods& Model relations as translation operations in the embedding space & TransE\cite{TransE}, TransH\cite{TransH}, TransR\cite{TransR}, RotatE\cite{RotatE} \cr
 \cmidrule{2-4}
 & GNN-based Methods& Use GNNs with aggregation operation to obtain the embeddings for reasoning & GCN\cite{GCN}, R-GCN\cite{RGCN}, CompGCN\cite{CompGCN} \cr
 \cmidrule{2-4}
 & LLM-based Methods& Transform KG into sequence, and adopt pre-trained LLMs for reasoning & KBERT\cite{KBERT}, CKBC\cite{CKBC}, KG-BERT\cite{KG-BERT}, StAR\cite{StAR}, MTL-KGC\cite{DBLP:conf/coling/KimHKS20} \cr
 \cmidrule{2-4}
 & Neural-symbolic & Combine the advantages of neural networks and symbolic logic reasoning & KALE\cite{KALE}, RUGE\cite{RUGE}, PRA\cite{PRA}, MINERVA\cite{MINERVA}, GraIL\cite{GraIL} \cr
\midrule

 \multirow{6}{*}{Dynamic KG}& 
 GNN-based Methods& Use GNN to capture inductive information and transfer it to new components  & DRUM~\cite{sadeghian2019drum}, TACT~\cite{TACT}, LAN~\cite{LAN}, MaKEr~\cite{MaKEr}, SE-GNN~\cite{SE-GNN} \cr
 \cmidrule{2-4}
 & LLM-based Methods& Convert DKG into sentences and obtain pre-trained prior knowledge from LLMs  & KEPLER~\cite{KEPLER}, BERTRL\cite{BERTRL}, Bi-Link~\cite{Bi-Link}   \cr
 \cmidrule{2-4}
 & Few-shot Methods& Transfer and adapt learned knowledge from various views to few-shot entities & FAAN~\cite{FAAN},  MetaR~\cite{MetaR}, Meta-KGR~\cite{Meta-KGR}, GEN~\cite{GEN}, REFORM~\cite{REFORM}  \cr

\midrule
 \multirow{16}{*}{Temporal KG}& Tensor Decomposition-based Completion Methods & The temporal information is regarded as one dimension of the TKG high-dimensional tensor, and the  temporal representation is obtained through tensor decomposition & TComplEX~\cite{lacroix2020tensor}, ~\cite{tresp2017embedding}, ConT~\cite{ma2019embedding},~\cite{shao2022tucker}, TIMEPLEX~\cite{TIMEPLEX}, TeLM~\cite{TeLM}, TGEOME~\cite{TGeomE}  \cr
 \cmidrule{2-4}
 & Translation-based Completion Methods & Temporal information is independent of the representation of entities and relationships and is explicitly represented as trainable vectors or projection matrices & TAE\cite{TAE2016}, TTransE~\cite{leblay}, HyTE\cite{HyTE}, TeRo~\cite{TERO}, TOKEI~\cite{leblay2020towards}, TA-TRANSE~\cite{TA-TransE}, TRESCAL~\cite{leblay}, DE~\cite{DE-SimplE}, ATiSE~\cite{ATiSE}  \cr
  \cmidrule{2-4}
 & Path-based Prediction Methods & The acquisition of temporal walks indicates the evolution law of TKG & TLogic ~\cite{TLogic}, TITer~\cite{sun2021timetraveler}, CluSTeR~\cite{li2021search}, xERTE~\cite{han2021explainable}  \cr
  \cmidrule{2-4}
 & Temporal Point Process-based Prediction Methods & The temporal knowledge as asynchronous events in a continuous time domain, and model the occurring probability of events  through temporal point process & Know-evolve~\cite{Know-evolve}, GHNN~\cite{han2020graph}, DyRep~\cite{trivedi2018dyrep}, M$^2$DNE~\cite{lu2019temporal}, TDIG~\cite{TDIG}, EvoKG~\cite{EvoKG}  \cr
  \cmidrule{2-4}
  & Sequence Neural Network-based Prediction Methods & The snapshot encoder outputs the representation of different time entities and relations, and the temporal encoder learns the evolution law in the TKG & RE\_NET~\cite{jin2019recurrent}, RE-GCN~\cite{RE-GCN}, CEN~\cite{CEN}, TeMP~\cite{TeMP}, DySAT~\cite{DGNN5}, DANE~\cite{xu2020embedding}, DACHA~\cite{chen2021dacha}, HIP~\cite{he2021hip} \cr
\bottomrule
\end{tabular}
\end{center}
\end{table*}

\textit{Self-attention-based Temporal Encoding Methods.} The self-attention mechanism can effectively alleviate the forgetting problem in encoding long-distance time series. TeMP \cite{TeMP} uses R-GCN\cite{RGCN} as the snapshot encoder to extract the historical vector sequence of the target entity s, $x_{s,t-\tau},\ldots,x_{s,t-1},x_{s,t}$; based on the self-attention mechanism, the temporal encoder aggregates the historical vector sequence with weights to obtain the vector representation $z_{s,t}$ of entity s at time t. 
Furthermore, DySAT\cite{DGNN5}  proposes a structural-temporal attention mechanism. The structure self-attention mechanism extracts features from the entity neighborhoods in each snapshot and aggregates them with weights, while the temporal self-attention mechanism captures the evolution law by flexibly weighting the historical entity representations of multiple time steps. Moreover, DANE\cite{xu2020embedding}, DACHA\cite{chen2021dacha}, HIP\cite{he2021hip} also design structure-temporal self-attention mechanisms, which are applied to the entity and relation prediction tasks.

\begin{figure*}
    \centering
    \includegraphics[width=\textwidth]{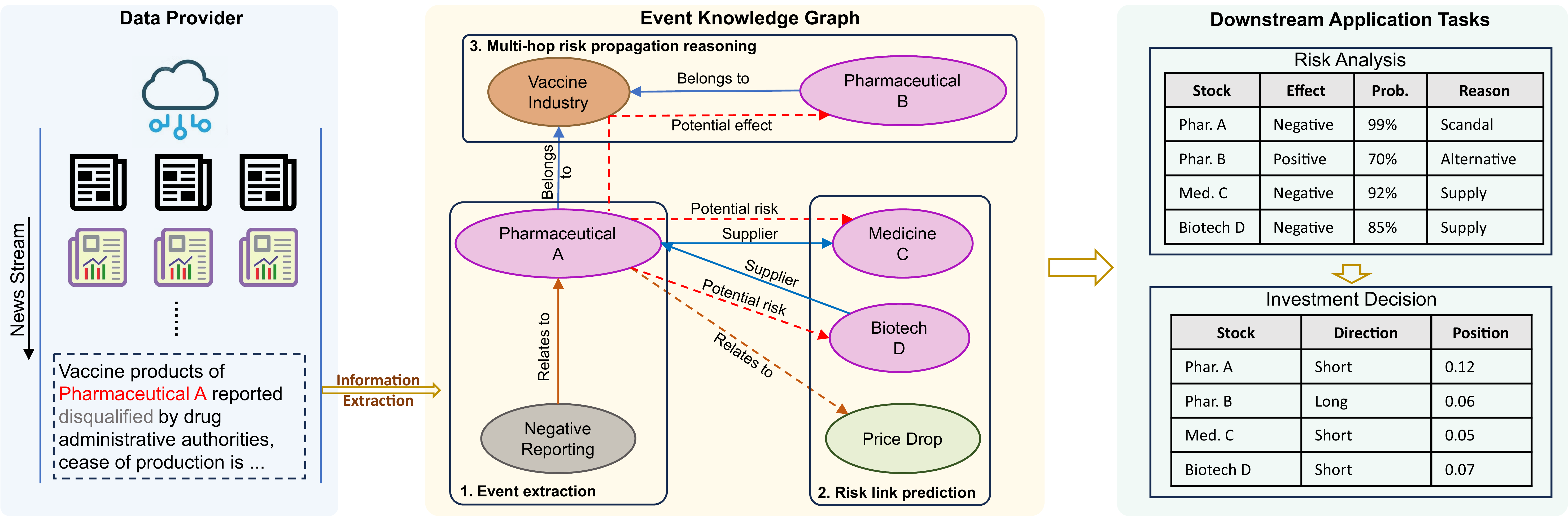}
    \caption{In an application of EKGR in event-driven quantitative investment, a piece of news concerning "Pharmaceutical A" (Phar. A) triggers the extraction and mapping of the event triple ("negative report", "relates to", "Phar. A") onto the EKG. Subsequently, link prediction is used to assess potential outcomes, such as the link ("Price drop", "relates to", "Phar. A"). Risk links, denoted as ("Phar. A", "Potential risk", $\ast$), are predicted for neighboring entities within the EKG, including upstream and downstream partners in the supply chain. Multi-hop reasoning is applied to analyze risk propagation among stocks within the same industry. These reasoning results are then utilized in risk analysis and investment decisions. For instance, negative effects propagating through Phar. A's supply chain may suggest short positions ("Med. C" and "Biotech D"), while long positions could be considered for companies offering alternative products ("Phar. B"). This illustration's depicted news contents, company names, and numerical values are purely fictional and solely serve illustrative purposes.}
    \label{fig:ekgr_example}
\end{figure*}

\section{Applications}~\label{sec:5}
Knowledge graphs have become integral parts of several areas of technology, from structured data management to artificial intelligence. They encapsulate structured information in a way that both humans and machines can comprehend, enabling a multitude of practical applications. Their seamless integration with LLMs has expanded these applications even further, making them indispensable in the landscape of data and AI. 

This section will first take the financial as a case to illustrate the applications of KG, and highlight the main application domains of KGs.

\subsection{Application in Finance}
The application of knowledge graphs in the finance sector serves as an exemplary illustration of how structured knowledge representation can transform data-driven industries. Specifically, knowledge extraction, knowledge reasoning, and intelligent QA systems that combine KGs with LLMs play a vital role in this domain.

\subsubsection{Financial Information Extraction and Knowledge Graph Construction}
The first essential step involves extracting information from financial documents, such as financial reports, economic news, and research papers, to construct or update a financial knowledge graph. In this context, information extraction targets various elements, including financial entities (e.g., stocks, bonds, corporations), relatively stable concepts (e.g., market indices, financial instruments), dynamically evolving knowledge (e.g., stock prices, economic indicators), temporal information (e.g., trade dates), and financial events (e.g., mergers, acquisitions, dividends announcements).

The extracted knowledge is then organized into a KG. This process transforms massive amounts of unstructured financial data into high-quality, structured financial knowledge, which is then effectively represented and made easily accessible through the KG.

\subsubsection{Financial Knowledge Reasoning}
Upon constructing the financial knowledge graph, graph-based data mining algorithms can be applied to infer new knowledge. For instance, the graph can be analyzed to detect patterns and trends based on historical financial events. One practical application of this is event-driven quantitative investment, where investment strategies are formulated based on the relationships and patterns identified within the KGs.

Figure~\ref{fig:ekgr_example} showcases an instance of EKGR application within the context of event-driven quantitative investment. This example demonstrates how investment decisions can be based on streaming events associated with entities in the financial market. Real-time news and reports serve as the data source, feeding relevant events into the investment system. Information in these events is then extracted and mapped to triples on the event knowledge graph. Subsequently, link prediction techniques are employed to analyze the potential impact of these events. By examining the links between events and associated stocks, the system can predict outcomes and assess risks. Additionally, multi-hop reasoning enables the analysis of risk propagation, particularly among stocks within the same industry. The results of this reasoning process are valuable for downstream applications, including risk analysis, stock trend prediction, and informing investment decisions such as determining long or short positions on specific stocks.

\subsubsection{Integration with LLMs for Financial QA}
Financial knowledge graphs can also be integrated with large language models to enable intelligent QA systems tailored for the finance sector. Through dynamically updated financial knowledge graphs, LLMs can be kept current with real-time financial knowledge. This integration not only enhances the ability of LLMs to perform tasks in the financial domain but also allows for fact-consistency checks, ensuring the factual accuracy of the content generated by LLMs.

Furthermore, the interaction with LLMs becomes more engaging and informative for users. For instance, a user can query the intelligent QA system regarding investment strategies based on recent financial events, and the system can provide insightful responses grounded in the data represented within the knowledge graph.

In conclusion, the finance sector benefits immensely from knowledge graphs by streamlining information extraction, enabling sophisticated knowledge reasoning, and enhancing user interaction through the integration of large language models. This amalgamation of technologies paves the way for more informed decision-making and a deeper understanding of complex financial ecosystems.

\subsection{Other Applications}
Knowledge graphs have widespread applications beyond finance, notably in healthcare, transportation, retail, manufacturing, legal, and personalized services.

\textbf{Healthcare}.
Knowledge graphs amalgamate patient records, medical literature, and drug information, paving the way for enhanced diagnosis, smart treatments, and drug discovery by elucidating relationships between diseases, symptoms, medications, and patient histories.

\textbf{Transportation}.
In transportation, knowledge graphs structure data encompassing routes, traffic conditions, and vehicle statuses, empowering intelligent systems to optimize route planning, manage traffic efficiently, and predict maintenance needs.

\textbf{Manufacturing}.
Within manufacturing, knowledge graphs assimilate data on product components, manufacturing processes, and supply chains, supporting predictive maintenance, quality control, and optimized production planning by analyzing relationships between machine performances, material availability, and production schedules.

\textbf{Legal}.
In the legal domain, knowledge graphs consolidate statutes, case law, and legal doctrines, aiding in legal research, case preparation, and regulatory compliance by structuring and analyzing the relationships between different legal entities and concepts.

\textbf{Personalized Services}.
Through social knowledge graphs modeling users' cognition and preferences, personalized services have burgeoned across sectors like education, psychotherapy, and personalized recommendations. These graphs facilitate a tailored user experience by aligning services and content with individual user preferences and behaviors, thereby enhancing engagement and satisfaction.

In summation, KGs significantly contribute to data organization and analysis across disparate sectors. The interplay between KGs and advanced technologies like LLMs further amplifies their application spectrum, propelling innovation and efficiency.

\section{The Future Directions of Knowledge Engineering}~\label{sec:6}
Existing researches on knowledge extraction, reasoning, and augmentation with LLMs for knowledge graphs have made significant progress in various aspects. However, as the forms of KGs have evolving, the corresponding techniques still face various challenges and open problems in the future. In this section, we try to point out promising research directions for future work.

\subsection{On the Evolution of Knowledge Representation}
The evolution of KGs illuminates the gradual understanding and also the limitations of this knowledge representation paradigm. Current challenges faced by KGs include: (1) the inability to directly and accurately model higher-order logic knowledge and apply it in higher-order knowledge reasoning; (2) ineffective handling and direct computation of numerical knowledge through KGs; (3) from a knowledge management perspective, the demand for a representation that supports continuous business iteration while effectively avoiding combinatorial explosion and redundant construction, challenges the existing KGs in unifying different granularities and types of knowledge and modeling their interactions, with RDF/OWL-based knowledge modeling being overly complex and prone to redundancy in rapidly evolving scenarios; and (4) unlike LLMs that store and reason knowledge through parameterization, the segregation of knowledge storage and reasoning in existing KGs lacks a unified framework integrating large-scale knowledge storage with efficient knowledge reasoning, hindering the maximization of knowledge utilization.

To address these challenges, exploring innovative knowledge representation paradigms is crucial. The notion of knowledge programmability, representing knowledge through programming languages, emerges as a potential avenue warranting further exploration. This proposition aligns with the broader goal of enhancing the granularity and expressiveness of knowledge representation, fostering a more nuanced understanding and management of the evolving knowledge landscape. By leveraging the latest advancements in knowledge representation technology, especially within the context of artificial intelligence and machine learning, both the academic and industrial sectors are poised to unveil new dimensions of knowledge representation transcending the traditional boundaries of KGs, paving the way for more dynamic, scalable, and comprehensive knowledge engineering systems.

\subsection{On the Evolution of Knowledge Extraction}
The construction of KGs relies on various knowledge extraction techniques. Most of the existing knowledge extraction methods are supervised models needing a large amount of labeled data. In order to effectively reduce the construction cost of KGs, it is necessary to go into zero-shot/unsupervised knowledge extraction methods and knowledge graph maintenance.

\subsubsection{From Supervised to Zero-shot/ Unsupervised Knowledge Extraction}
Most supervised knowledge extraction methods are based on contextualized LLMs fine-tuned by labeled data and corresponding tasks. However, these labeled data often need to be annotated by humans, especially for some knowledge-intensive texts, the annotation process requires certain domain knowledge and could be very costly. Meanwhile, generative LLM models have been viewed as a kind of strong zero-shot learner that can be applied to various types of downstream tasks in scenarios where labeled data is lacking. Nevertheless, the existing generative LLMs still significantly underperform fully-supervised baseline models on knowledge extraction tasks. This is because most of the knowledge extraction tasks are labeling tasks, while generative LLMs are good at generation tasks. To achieve zero-shot or unsupervised knowledge extraction for low-cost knowledge graph construction, it is essential future direction to fill this gap by efficiently transforming different types of knowledge extraction tasks into generation tasks and solving them using generative LLMs.

\subsubsection{Potential Direction of Knowledge Quality Control and Maintenance}
Quality control and maintenance are crucial in knowledge engineering, affecting KG usability in real-world applications because it is hard to ensure the completeness and correctness of extracted knowledge. The advent of LLMs provides new perspectives on addressing these challenges, which leverages parameterized knowledge of LLMs learned from vast corpora for KGs' quality control and maintenance in an unsupervised way.

\subsection{On the Evolution of Knowledge Reasoning}
The existing knowledge reasoning techniques mainly focus on SKGs. With the formal evolution of KGs, knowledge reasoning techniques for DKGs and TKGs have received more attention, but still remain to be improved. It is also worth noting that reasoning over EKGs is largely unexplored. Besides, neural-symbolic knowledge reasoning also remains to be explored.

\subsubsection{Non-static Knowledge Graph Reasoning}
Recent advances have prompted a shift from SKGs to more dynamic structures like DKGs and TKGs, aiming to accurately represent and reason over the dynamic nature of real-world knowledge. However, reasoning techniques for DKGs and TKGs are still nascent, with temporal knowledge completion and extrapolation emerging as notable endeavors. The development of techniques to navigate the temporal and dynamic dimensions of knowledge is crucial, presenting a fertile ground for future research.

A significant avenue for future research is to unravel the evolution logic behind temporal knowledge, promising to enrich the reasoning frameworks for these non-static KGs. This includes developing advanced algorithms for tracking and modeling temporal evolution of facts and relationships, and exploring how different types of temporal knowledge, such as dynamic relations and interactions, interconnect and influence each other.

\subsubsection{Event Knowledge Graph Reasoning} 
In recent years, research on knowledge reasoning in SKGs, DKGs, and TKGs has attracted widespread interest. With the rise of EKGs, EKG reasoning (EKGR) will inevitably become a new research hotspot. EKGR could probably predict potential future events based on the existing relations between entities and events, which is very important for many practical application scenarios. However, how to develop an EKG embedding model to model events, a new kind of knowledge element, has never been discussed in previous research. It will also be interesting to explore the utility of event logic and LLMs for the task of EKGR since EKGs contain a lot of logical relationships and textual descriptions between events.

Despite the promising value of EKGR in practical applications, several challenges need to be addressed to enable its effective implementation in the real world. Firstly, the development of a robust EKG embedding model is crucial as it forms the foundational platform for accurate EKG reasoning. Secondly, suitable data structures and storage techniques must be employed to handle the continuously growing influx of event information. Adequate accommodations are needed to ensure efficient storage and retrieval of large volumes of event data. Lastly, the real-time nature of many applications necessitates the resolution of algorithmic and engineering challenges to improve the execution speed of EKGR algorithms.

\subsubsection{From Neural or Symbolic to Neural-Symbolic Knowledge Reasoning}
Recent advancements in neural or symbolic reasoning have highlighted the potential of integrating neural networks' data-driven learning with symbolic reasoning's structured logic, to enhance reasoning on various KGs. This fusion aims to overcome limitations posed by exclusively neural or symbolic approaches, particularly in representing the dynamic and time-variant nature of real-world knowledge in DKGs and TKGs.

Besides factual knowledge, a considerable amount of logical knowledge is implied in KGs, encompassing a vast number of first-order or temporal logical rules between relations and events. The application of these logical rules, when combined with various KG embedding models, can significantly improve the performance and interpretability of these models. Some existing works have demonstrated substantial improvements in static knowledge graph reasoning tasks by applying first-order logical rules. Yet, the application of static and temporal logical rules to KG reasoning tasks remains an open avenue for exploration. Furthermore, the emergence of LLMs presents an opportunity to uncover underlying patterns in knowledge, aiding in reasoning tasks. The blend of neural-symbolic paradigms and the pattern recognition capabilities of LLMs could potentially pave the way for more effective and interpretable KG reasoning techniques, tailored to the evolving nature of knowledge in various types of KGs.

\subsection{Potential Combinations of KGs and LLMs}
Knowledge Graphs and Large Language Models represent contrasting paradigms in knowledge representation. KGs are the most representative symbolic knowledge form, which provides interpretability through structured information, yet may lack flexibility.
LLMs are the most powerful parameter knowledge form, which excels in generativity and adaptability, handling diverse tasks, but can be opaque in decision-making.
To gain a deeper understanding of the KG and LLM's similarities and differences, and guide the future integration and application of both, we propose a comprehensive comparison.

\subsubsection{Similarities}\label{sec:5.1}
The main similarities between KGs and LLMs are three-fold: 
(1) both of their data sources come from massive amounts of unstructured data; 
(2) both of them can represent entity semantics, and store correlations between entities; 
(3) both of them can be used as knowledge sources for downstream applications, e.g., question answering.

The training of LLMs requires large amounts of textual data, which are often obtained from publicly available Internet websites. The construction of large-scale KGs also relies on information extraction from unstructured open data. Some LLMs and large-scale KGs even use the same data sources. For example, Wikipedia is the most important training corpus for a lot of LLMs like BERT, and also the main data source for general KGs, e.g., DBpedia, and YAGO.

KGs express the semantic information of entities by recording their types, attributes, and descriptions, as well as relations between them. On the other hand, LLMs are trained on text data to obtain character-level embeddings or word-level embeddings of entities, which can also represent the semantics of entities. LLMs~\cite{KGLM2} have also been shown to be able to recall relational knowledge without fine-tuning. Specifically, to answer a query (\textit{Joe Biden}, \textit{place\_of\_birth}, \textit{?}) using an LLM, one can input a masking sentence \textit{"Joe Biden was born in [MASK]"} or a natural language generation (NLG) task \textit{"Where was Joe Biden born? [MASK]"} to probe relational knowledge, where \textit{MASK} is a placeholder for the object entity to be predicted and \textit{"was born in"} is the prompt string of relation \textit{"place\_of\_birth"}. Previous studies have shown that such paradigms can perform well on some knowledge probing benchmarks like LAMA~\cite{KGLM1}. Moreover, some follow-up works propose various automatic prompt generation methods to further improve probing performances of LLMs~\cite{knowprob1,knowprob2,knowprob3,knowprob4,knowprob5}.

Large-scale KGs are often the main knowledge sources of some downstream applications, e.g., a knowledge base question answering (KBQA) system. The rise of LLMs makes it possible to perform such downstream tasks without access to external knowledge, e.g., closed-book QA. For instance, BERT has been proven to do remarkably well on open-domain QA~\cite{KGLM1}. Some literature shows that such LLMs can further improve their performances on open-domain QA after fine-tuning~\cite{LMQA}. 

\subsubsection{Differences}\label{sec:5.2}
To illustrate the differences between LLMs and KGs, we discuss the limitations of KGs with respect to LLMs and the current limitations of LLMs relative to the capabilities of KGs, separately. We conclude the limitations of KGs as follows:

\textbf{(1) High construction cost:} The construction of knowledge graphs requires complex NLP techniques to process the text and quality testing to ensure the accuracy of the knowledge. Thus, the construction and maintenance of a large-scale KG often need significant human effort.

\textbf{(2) Data Sparsity:} Due to the high labor cost, the domain and data volume covered by a KG are limited, which leads to a large amount of missing data in the KG.

\textbf{(3) Lack of flexibility:} The storage structure and query method of a KG are relatively fixed, and it is difficult to adapt to various data structures and query requirements. By contrast, LLMs can offer better extendability and expressibility to different kinds of downstream tasks and thus have a broader range of application scenarios.

Although LLMs demonstrate superior reasoning capabilities and versatility over KGs, there are still some obvious limitations of LLMs that can be addressed by combining them with different types of KGs,

\textbf{LLM vs SKG:} LLMs are black-box models in nature. It is difficult to know the basis of the answers given by an LLM and the corresponding reasoning process. Meanwhile, one of the main advantages of KGs is their explainability. It would be easier to know how the answers are inferred over a KG. Moreover, regardless of knowledge updating and revision, an SKG will give a consistent answer to a query no matter how many times this query is performed. By contrast, LLMs might show different answers to the same question due to their randomness and nature of probabilistic models. And the training corpus used for LLM might be noisy. The training processes of LLMs have the risk of learning incorrect or toxic knowledge~\cite{LMdanger}, while the \textbf{reliability of knowledge} is ensured by quality assessment for a KG, especially for small-scale domain-specific KGs.

\textbf{LLM vs DKG:} LLMs might suffer from \textbf{outdated knowledge and information} since they are pre-trained based on Internet data at a specific time. And LLMs have \textbf{limited editability}, that is to say it is difficult for LLMs to frequently learn new knowledge and information due to their \textbf{unaffordable training cost}. For example, the largest GPT3 model~\cite{gpt3} has a parametric count of 175B and costs 10 million dollars for a single training session. By contrast, it is much easier for DKGs to revise and update the knowledge they have stored. Thus, the cost of knowledge updating will be significantly reduced for LLMs if LLMs and DKGs can be effectively combined without expensive training processes.

\textbf{LLM vs TKG:} LLMs face three main challenges when dealing with temporal information~\cite{TKGLM}. The first challenge is averaging, i.e., the model may see conflicting information about time ranges. The second challenge is forgetting, where the model may not remember facts that are valid only for a representative period of time, and therefore performs poorly when asked questions about the more distant past. The third challenge is poor temporal calibration, i.e., as LLMs become "stale", they are increasingly likely to be queried for facts that are outside the time range of their training data. Relatively, TKGs tend to have longer time scopes and the factual knowledge in TKGs is much more time-sensitive.

\textbf{LLM vs EKG:} LLMs have achieved noticeable success on many NLP tasks, but they still struggle with event-centric tasks. One of the main disadvantages of LLMs in learning event-centric knowledge is that they may not be able to capture the temporal relations between events effectively. This is because LLMs are trained on large text corpora via self-supervised learning methods and may not have access to structured data that can help them learn about event temporal relations~\cite{econet}. Moreover, event-centric knowledge is updated very quickly, and it is difficult for LLMs to learn from the constant addition of new event-centric knowledge.  

In addition to their respective limitations, KGs and LLMs have different representation forms to store knowledge and different ways of performing knowledge reasoning. And although LLMs obtain decent knowledge probing performances on LAMA, some follow-up works propose more difficult and domain-specific probing datasets~\cite{BioLAMA, MedLAMA} for LLMs and demonstrate that current LLMs are not reliable in factual knowledge extraction~\cite{KGLM3, EBERT}, which is contrary to the perspective that LLMs can be seen as knowledge bases. Therefore, it is necessary to address the existing limitations of LLMs to combine various KGs.

\subsubsection{The Combination of LLMs and Different KGs}\label{sec:5.2}
\begin{figure}[t]
\centering
\includegraphics[width=0.8\linewidth]{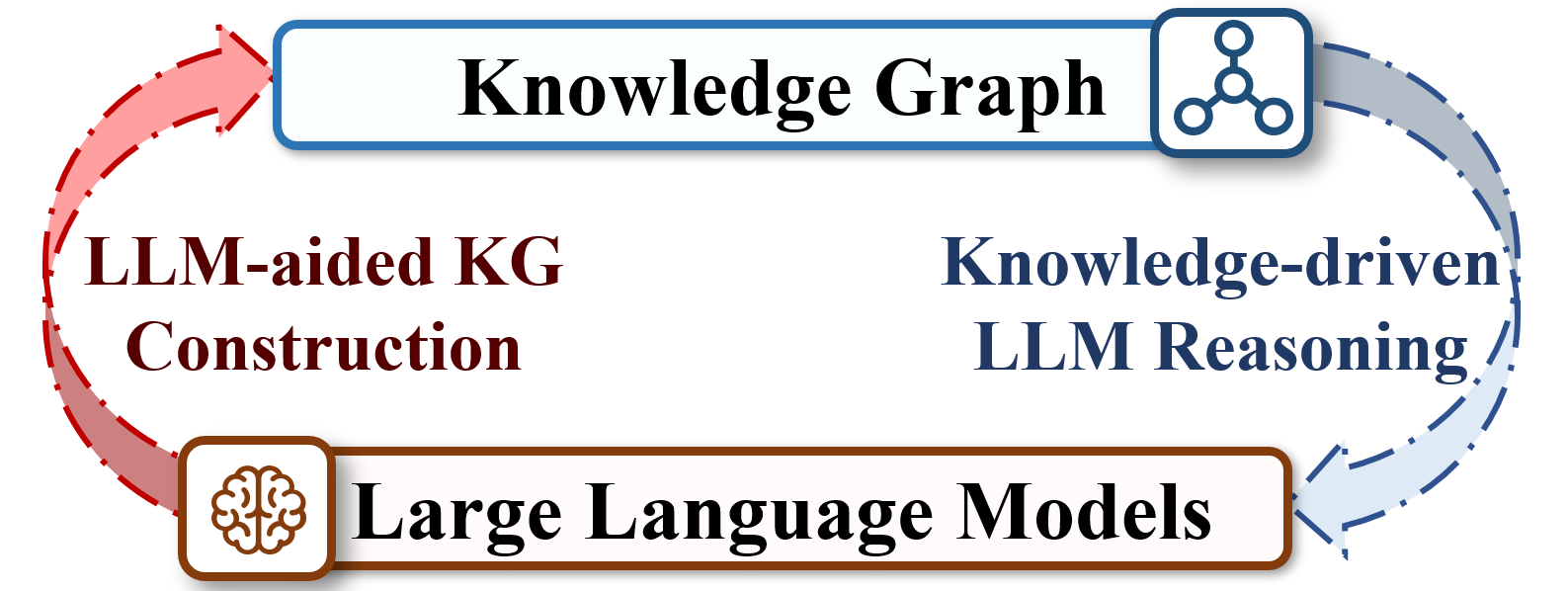} 
\caption{The potential integration framework of KGs and LLMs.}
\label{fig:KGLLM}
\end{figure}

In summary, despite the distinct characteristics and functionalities of KGs and LLMs, there is significant potential for integration between the two, as illustrated in Figure~\ref{fig:KGLLM}.

Specifically, from the perspective of \textbf{LLM aided KG construction}: Utilizing the text processing capabilities of LLMs, KGs can achieve more efficient unsupervised knowledge extraction during their construction phase, thereby enhancing their automated building and editing capabilities. Additionally, LLMs can offer unsupervised quality control and maintenance for KGs, ensuring the quality and accuracy of knowledge. In terms of knowledge inference, the pattern induction capabilities of LLMs present new opportunities for neural-symbolic reasoning.

From the perspective of knowledge-driven LLM reasoning, KGs serve as a repository containing both logical and factual knowledge, providing a richer source of information. During the training phase, future research will increasingly explore how to integrate structured, high-quality knowledge with LLMs, such as using graph neural networks for knowledge encoding or adopting new encoding strategies. Such methods not only enrich the model's knowledge base but also optimize its structure and training objectives. In the inference phase, integrating KGs to detect and reduce knowledge illusions and enhance the reliability and accuracy of model outputs will be a pivotal research direction~\cite{sun2023think, wen2023mindmap, ma2024think}.

In conclusion, the integration of KGs and LLMs will bring innovation to the AI domain. By complementing each other, both entities have the potential to achieve superior performance and broader application scenarios.

\section{Conclusion}
Since the emergence of the technology of KG, the form of KGs is gradually changing. KG-related techniques, like knowledge extraction and reasoning, have also evolved with the change in the form of KGs.
In this survey, we first provide a comprehensive overview of the evolution of KGs, from static to dynamic, temporal, and event-based KGs, and give formal definitions specific to different types of KGs.
In addition, we survey the existing knowledge extraction and knowledge reasoning methods and classify these methods according to the types of target KGs. Moreover, we discuss the applications together with a case study of the financial and other scenarios. Finally, promising future research directions about knowledge engineering are clarified further based on the developmental tendencies of the existing work.


\section*{Acknowledgment}
This work is supported by the National Natural Science Foundation of China (No.U1836206, 91646120, U21B2046, 62172393), the National Key Research and Development Program of China under grants (No. 2018YFB1402601), the Zhongyuanyingcai program-funded to central plains science and technology innovation leading talent program (No.204200510002) and Major Public Welfare Project of Henan Province (No.201300311200).


%

\appendices




\ifCLASSOPTIONcaptionsoff
  \newpage
\fi



%

%

\bibliographystyle{IEEEtran}
\bibliography{ref_sim}







\end{document}